\RequirePackage{etoolbox}
\RequirePackage[table,dvipsnames]{xcolor}

\newtoggle{neurips} %
\toggletrue{neurips} %

\documentclass{article}
\PassOptionsToPackage{numbers, compress}{natbib}
\usepackage{neurips_data_2021}

\usepackage[utf8]{inputenc} %
\usepackage[T1]{fontenc}    %
\usepackage{hyperref}       %
\usepackage{url}            %
\usepackage{booktabs}       %
\usepackage{amsfonts}       %
\usepackage{nicefrac}       %
\usepackage{microtype}      %
\usepackage{times}
\usepackage{epsfig}
\usepackage{graphicx}
\usepackage{amsmath}
\usepackage{amssymb}
\usepackage{setspace}
\usepackage{bm}
\usepackage[accsupp]{axessibility}
\usepackage{pifont}%
\newcommand{\cmark}{\ding{51}}%

\iftoggle{neurips}{
}{
\papertype{Original Article}
\paperfield{Journal Section}
}

\usepackage{siunitx}
\usepackage{mysymbols}
\usepackage{multirow}
\usepackage{makecell}
\usepackage{rotating}

\title{Neural Latents Benchmark `21: Evaluating latent variable models of neural population activity}

\iftoggle{neurips}{
\author{%
    Felix Pei$^{1}$\thanks{Equal contribution.}\;,
    Joel Ye$^{1,2*}$,
    David Zoltowski$^{4}$,
    Anqi Wu$^{1,5}$,
    \\
    \textbf{Raeed H. Chowdhury$^{6}$,
    Hansem Sohn$^{7}$,
    Joseph E. O'Doherty$^{8}$,
    Krishna V. Shenoy$^{9}$},
    \\
    \textbf{Matthew T. Kaufman$^{10}$,
    Mark Churchland$^{5}$,
    Mehrdad Jazayeri$^{7}$,
    Lee E. Miller$^{11}$},
    \\
    \textbf{Jonathan Pillow$^{4}$,
    Il Memming Park$^{12}$,
    Eva L. Dyer$^{1,3}$,
    Chethan Pandarinath$^{1,3}$\thanks{Correspondence to \texttt{chethan@gatech.edu}}}\quad    \vspace{2mm} \\
 
    $^1$Georgia Institute of Technology,%
    $^2$Carnegie Mellon University,
    $^3$Emory University, \\
    $^4$Princeton University,%
    $^5$Columbia University,%
    $^6$University of Pittsburgh,\\
    $^7$Massachusetts Institute of Technology,%
    $^8$Neuralink Corp.,%
    $^9$Stanford University,\\
    $^{10}$University of Chicago,%
    $^{11}$Northwestern University, %
    $^{12}$Stony Brook University
}
}{
\author[1\authfn{1}]{Felix Pei}
\author[2,3\authfn{1}]{Joel Ye}
\author[6]{David Zoltowski}
\author[8]{Anqi Wu}
\author[9,10]{Raeed H. Chowdhury}
\author[11]{Hansem Sohn}
\author[12]{Joseph E. O'Doherty}
\author[13]{Krishna V. Shenoy}
\author[14]{Matthew T. Kaufman}
\author[15]{Mark Churchland}
\author[11]{Mehrdad Jazayeri}
\author[16]{Lee E. Miller}
\author[17]{Il Memming Park}
\author[1,4]{Eva Dyer}
\author[6,7]{Jonathan Pillow}
\author[4]{Chethan Pandarinath}

\contrib[\authfn{1}]{Equally contributing authors.}

\affil[1]{School of Electrical and Computer Engineering, Georgia Institute of Technology, Atlanta, GA, USA}

\corraddress{Chethan Pandarinath, Wallace H. Coulter Department of Biomedical Engineering and Department of Neurosurgery, Emory University and Georgia Institute of Technology, Atlanta, GA, USA}
\corremail{chethan@gatech.edu}

\fundinginfo{Funder One, Funder One Department, Grant/Award Number: 123456, 123457 and 123458; Funder Two, Funder Two Department, Grant/Award Number: 123459}

\runningauthor{Felix Pei et al.}
}

\begin{document}

\bibliographystyle{unsrt}

\newcommand{\maze}{\texttt{MC\_Maze}~}
\newcommand{\mazeL}{\texttt{MC\_Maze-L}~}
\newcommand{\mazeM}{\texttt{MC\_Maze-M}~}
\newcommand{\mazeS}{\texttt{MC\_Maze-S}~}
\newcommand{\dmfc}{\texttt{DMFC\_RSG}~}
\newcommand{\rtt}{\texttt{MC\_RTT}~}
\newcommand{\areatwo}{\texttt{Area2\_Bump}~}

\iftoggle{neurips}{}{}

\maketitle

\begin{abstract}
Advances in neural recording present increasing opportunities to study neural activity in unprecedented detail. Latent variable models (LVMs) are promising tools for analyzing this rich activity across diverse neural systems and behaviors, as LVMs do not depend on known relationships between the activity and external experimental variables. 
However, progress with LVMs for neuronal population activity is currently impeded by a lack of standardization, resulting in methods being developed and compared in an ad hoc manner. 
To coordinate these modeling efforts, we introduce a benchmark suite for latent variable modeling of neural population activity. 
We curate four datasets of neural spiking activity from cognitive, sensory, and motor areas to promote models that apply to the wide variety of activity seen across these areas. 
We identify unsupervised evaluation as a common framework for evaluating models across datasets, and apply several baselines that 
demonstrate benchmark diversity.
We release this benchmark through EvalAI. 
\href{http://neurallatents.github.io}{http://neurallatents.github.io}

\end{abstract}

\section{Introduction}
A central pursuit of neuroscience is to understand how the rich sensory, motor, and cognitive functions of the brain arise from the collective activity of populations of neurons. To this end, we are witnessing a sea change in systems neuroscience, as a decade of rapid progress in neural interfacing technologies has begun to enable access to the simultaneous activity of vast neuronal populations \cite{stevenson2011advances}.
As a result, neuroscientists increasingly capture high-dimensional and dynamic portraits of activity from a variety of brain areas and during diverse behaviors. 
The resulting datasets may stymie traditional analytical approaches that were designed around recordings from one or a handful of neurons at a time.

In response to the increased data complexity, computational neuroscientists are producing powerful methods for uncovering and interpreting structure from neural population activity. An emerging and particularly promising set of approaches -- termed \emph{latent variable models} (LVMs) -- characterizes patterns of covariation across a neuronal population to reveal its internal state~\cite{hurwitz2021building}. LVMs have proven useful for summarizing and visualizing population activity, relating activity to behavior, and interrogating the dynamic mechanisms that mediate population-level computations~\cite{cunningham2014dimensionality,pandarinath2018latent,vyas2020computation,keeley2020modeling,dabagia2020comparing}.

A key opportunity to advance the application of LVMs to neural data is to capitalize on the dramatic advances in machine learning over the past decade. Yet there is a high barrier to entry for ML experts to make an impact, likely stemming from the lack of standardized datasets and metrics for evaluating and comparing LVMs. To address this gap, we introduce the \textbf{Neural Latents Benchmark} (NLB), a series of benchmark suites that will enable standardized quantitative evaluation of LVMs on neural data. These suites will provide curated datasets, standardized APIs, and example codepacks. 

Here we present our first suite, \textbf{NLB `21}, which aims to broaden the potential applicability of LVM approaches by benchmarking unsupervised modeling on datasets from a variety of brain regions, behaviors, and dataset sizes. While LVMs are often developed and evaluated using data from a single brain region and behavior, activity from different regions or behaviors may have markedly different structure and thus present different modeling challenges~\cite{keshtkaran2021large}. NLB ‘21 provides curated neurophysiological datasets from monkeys that span motor, sensory, and cognitive brain regions, with behaviors that vary from pre-planned, stereotyped movements to those in which sensory input must be dynamically integrated and acted upon. Further, because neuroscientific datasets can vary substantially in size and different LVMs may be more or less data-efficient, we introduce data scaling benchmarks that evaluate LVM performance on datasets of different sizes. To achieve general evaluation of LVMs regardless of brain region, behavior, or dataset size, we adopt a standardized unsupervised evaluation metric known as \emph{co-smoothing}~\cite{Macke2011-ep}, which evaluates models based on their ability to predict held-out neuronal activity. We also provide secondary evaluation metrics whose applicability and utility vary across datasets. To ensure accessibility and standardization, we provide datasets in the Neurodata Without Borders format~\cite{teeters2015neurodata}, host the benchmarks and leaderboards on the EvalAI platform~\cite{yadav2019evalai}, and offer a code package that demonstrates data preprocessing and submission. To ensure consistent evaluation and to mitigate issues with overfitting or hyperparameter hacking, model predictions are evaluated against private data that are unavailable to developers.

This manuscript first motivates the broad use of LVMs for interpreting neuroscientific data. We then detail benchmark datasets and evaluation metrics, and provide examples of their application with a variety of current LVM approaches. We anticipate that this benchmark suite will provide valuable points of comparison for LVM developers and users, enabling the community to identify and build upon promising approaches.

\subsection{Scientific Motivation and Evaluation Philosophy}
LVMs are a powerful approach for characterizing the internal state of biological neural networks based on partial observations of the neuronal population.
In applications to neuronal population activity, LVMs are generative models that 
describe observed activity as a combination of latent variables, which are often fewer than the number of observed neurons and typically exhibit an orderly progression in time.
LVM approaches to neural data are grounded in empirical findings that neurons in large neuronal populations do not act independently, but rather exhibit coordinated fluctuations~\cite{Yu2009-qp,Jazayeri2017,gao2015simplicity}. We point the reader to a recent review~\citep{hurwitz2021building} for discussion on how LVMs can be used for neuroscientific insights.
In NLB `21, we focus on unsupervised LVMs that are not directly conditioned on measured external variables. Without a strict dependence on external variables, such models have broad applicability, including settings where we cannot observe or even know the relevant external variables that affect a neural population's response. 

To understand the utility of LVMs in probing neural circuits, a helpful analogy is the task of reverse engineering an artificial network that takes in a set of inputs, performs some computation, and produces a set of outputs.
Between input and output, the artificial network has a set of intermediate representations, and studying these representations provides insight into the computational strategy employed by the network.
Likewise, neural population recordings are observations of intermediate representations of a biological neural network that processes information and coordinates behavior. 

However, the task of understanding these representations may be considerably more challenging for neural population recordings than for artificial networks, as we typically have recordings from a limited number of neurons in one or a few brain regions, we may not know anatomical connectivity, and there may be many steps of processing between externally-measurable inputs and outputs and the brain area(s) under study. Further, the observed neuronal responses are highly variable and seemingly noisy~\cite{cunningham2014dimensionality}, making the task of understanding neural population activity inherently statistical and well-suited for latent variable approaches.

There are many potential ways to model neural population activity with latent variables, and different assumptions can lead to varied model structures that are all seemingly ``correct.''
Given this ill-posed nature of latent variable modeling, we sought a primary evaluation approach that was largely agnostic to the form and structure of the LVM being evaluated. Indeed, the co-smoothing evaluation (defined in Section~\ref{cosmooth}) is an unsupervised approach that only assesses the ability of LVMs to describe the observed neuronal activity itself. This allows for flexibility in modeling assumptions, while avoiding the intricate complexity of comparing vastly varying structures of LVMs.

\subsection{Related Work} 
\xhdr{Evaluation strategies for LVMs applied to neural data}
Many new LVMs have been developed to meet the need to analyze large scale neural population recordings. For context, we document appearances of such neural data LVMs in ML venues in supplementary~\tableref{tbl:lvm_index}. These LVMs are evaluated on a variety of datasets which collectively span many areas of the brain. Other animal models (rats~\citep{Wu2017-az}, mice~\citep{wu_olfactory_2018}), and non-electrophysiological recording modalities (calcium imaging~\citep{wu_olfactory_2018}, fMRI~\citep{cai_incorporating_2020}) are used as well. The diversity of previously used datasets is impressive, but the lack of standardized datasets has made comparison across models difficult. Moreover, even when two LVMs use the same dataset, they will often report on different metrics, or different variations of the same metric, as evidenced in~\tableref{tbl:lvm_index}. Non-standardized evaluation has made it extremely difficult to track the state of the field.

\xhdr{Existing benchmarks in computational neuroscience}.
The computational neuroscience community has recently produced several benchmarks around the interaction of machine learning methods and neural data.
Some focus on the challenge of extracting spiking activity (action potentials, or correlates) from raw neurophysiological data, for example, the spikefinder challenge~\cite{berens2018community} for inferring spiking activity from two-photon calcium imaging data, and SpikeForest~\cite{magland2020spikeforest} for extracting spikes from electrophysiological recordings. Other benchmarks evaluate single neuron modeling to predict spike times~\cite{Gerstner2009spkpred} and decoding externally-measurable variables from neural population activity in motor and somatosensory cortices and hippocampus~\cite{glaser2020machine}. Separately, Brain-Score evaluates the alignment of deep neural networks trained to perform behavioral tasks and the brain areas associated with those tasks (particularly the ventral visual stream)~\cite{schrimpf2020brain}. Distinct from all of these, NLB '21 quantifies how well LVMs can describe neuronal population activity in an unsupervised manner.

\xhdr{Evaluating generative model latents and outputs}.
A candidate approach for comparing LVMs is to evaluate the quality of the model's intermediate representations. In ML, such intermediate latent representations are typically assessed via supervised evaluation, \eg by measuring their disentanglement relative to data categories~\cite{higgins2016beta} or transfer performance in a variety of downstream tasks \cite{bengio2013representation}. As described earlier, however, supervised evaluation may be limiting in some neuroscience applications, where external variables may be unreliable or incomplete descriptions of neuronal activity (particularly in higher order brain areas, such as those underlying cognition). Therefore, achieving performance metrics that generalize across brain regions and task conditions requires unsupervised evaluation.

A common approach to unsupervised evaluation of latent representations is to assess the quality of a model's output, via one-to-one comparisons with reference data. This can be achieved via prediction of held-out portions of the data, as in MLM (masked language modeling)~\citep{devlin2019bert} or image inpainting~\citep{doersch2015unsupervised}. When used as a learning objective, these metrics can risk ``shortcut'' solutions based on shallow data correlations~\citep{doersch2015unsupervised}. On the other hand, MLM approaches have been demonstrably effective at inducing high-level, semantic representations~\citep{devlin2019bert}, justifying their use as a first proxy objective in absence of satisfactory supervised metrics.

\section{NLB '21: Datasets}

To facilitate the development of LVMs for broad application, our datasets span brain regions that underlie motor, sensory, and cognitive functions, and span a variety of behaviors. As described below, each dataset presents a unique set of challenges to uncovering neural latents. We note that while all datasets have been used in previous studies (primarily for neuroscientific purposes), there have not been any previous benchmarks for neural data LVM evaluation, either using these data, or any other that we know of.%

All datasets contain electrophysiological measurements recorded using intracortical microelectrode arrays. Preliminary signal processing was applied to the raw voltage recordings to extract spiking activity. While this process of spike sorting is imperfect and a subject of active study, we view it as a distinct problem from the process of extracting latent structure in the data, and thus provide datasets in spike-sorted form. Detailed descriptions of each dataset can be found in the Appendix. All datasets are available through DANDI (Distributed Archives for Neurophysiology Data Integration) in the NWB (Neurodata without Borders) standard~\cite{rubel2021neurodata}. Links can be found at \href{https://neurallatents.github.io/}{https://neurallatents.github.io/}. Note that our benchmark uses select recording sessions from larger datasets with potentially many other sessions. Thus, related data to those we present here may have been separately uploaded for public use (e.g., on DANDI), but those releases exclude the specific recording sessions used for NLB'21.

\paragraph{\maze\hspace{-1mm}.\hspace{-2mm}} The Maze datasets consist of recordings from primary motor and dorsal premotor cortices while a monkey made reaches with an instructed delay to visually presented targets while avoiding the boundaries of a virtual maze~\cite{churchland2010cortical}. The monkey made reaches in 108 behavioral task configurations, where each task configuration used a different combination of target position, numbers of virtual barriers, and barrier positions. These different configurations elicited a wide variety of straight and curved reach trajectories, and each configuration was attempted by the monkey many times in random order, resulting in thousands of trials recorded across a given experimental session.

\begin{figure}[t]
    \centering
    \includegraphics[width=1.0\textwidth]{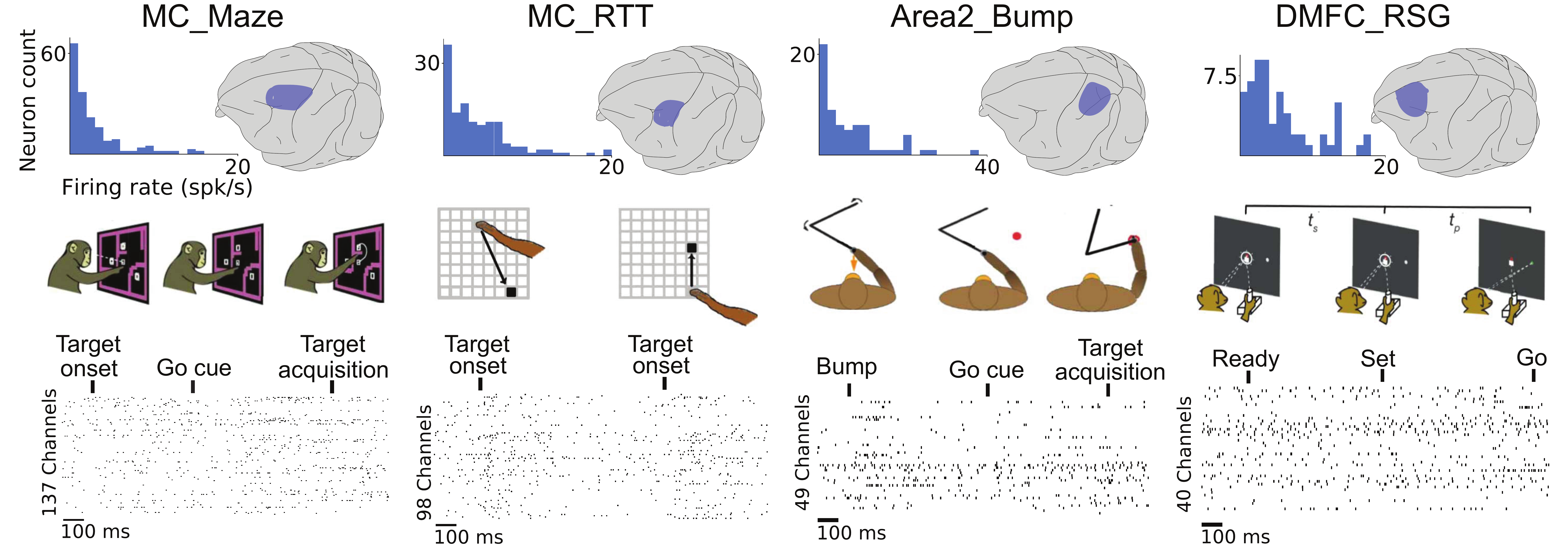}
\vspace{-3mm}
    \caption{\footnotesize\xhdr{NLB ‘21 datasets span four diverse brain area/task combinations}. For each behavioral task (center row), the top left panel presents the distribution of firing rates of the neurons in the dataset, while the top right highlights the recorded brain area. Lower panels present sample spike rasters, aligned to task events. \textbf{Tasks:}
    Motor cortical (MC) datasets include a center-out instructed delay reaching task with stereotyped conditions (Maze) and a continuous series of reaches in a random target task (RTT). Data from somatosensory cortex (Area 2) include externally-perturbed movements and volitional, goal-directed movements. Data from dorso-medial frontal cortex (DMFC) are during the Ready-Set-Go (RSG) cognitive time interval reproduction task.\vspace{-3mm}}
    \label{fig:datasets}
\end{figure}

The Maze datasets are exceptional in their combination of behavioral richness (number of task configurations), stereotyped behavior across repeated trials (tens of repeats for each task configuration), and high total trial counts (thousands) -- these attributes support averaging neuronal activity across repeated trials as a simple, first-pass de-noising strategy~\cite{cunningham2014dimensionality}, while retaining enough diversity in task conditions to allow rich investigation into the structure of the population activity~\cite{gao2017theory}. Additionally, the instructed delay paradigm allows movement preparation before presenting a go cue, which enables a clean separation of the neural processes related to preparation and execution~\cite{churchland2010cortical, shenoy2013cortical}. Due to the instructed delay paradigm and lack of unpredictable task events, population activity during the execution phase is largely predictable based on the state of the neural population at preparation, creating a unique case where activity can be well-modeled as an autonomous dynamical system~\cite{shenoy2013cortical,Churchland2012-JPCA,Pandarinath2018-LFADS,pandarinath2018latent}. With their unique properties, the Maze datasets have been extensively used in neuroscientific studies - in particular, they have been critical for revealing a plethora of insights into the structure of neural population activity during movement preparation and execution~\cite{churchland2010cortical,Churchland2012-JPCA,hennequin2014optimal,elsayed2017structure,kaufman2010roles,sussillo2015neural,kaufman2013roles,kaufman2014cortical,kaufman2016largest,kaufman2015vacillation}. They have also been used for validating a few LVMs individually~\cite{keshtkaran2019enabling,Pandarinath2018-LFADS,ye2021representation,keshtkaran2021large}.

\maze consists of one full session with 2869 total trials and 182 neurons, with simultaneously monitored hand kinematics. We expect this dataset to serve as a basic yet versatile baseline for LVM development (akin to a ``neuroscience MNIST''). Additionally, to support a data-scaling benchmark that characterizes data efficiency of LVMs, we provide three scaled datasets, each from separate experimental sessions, containing 500, 250, and 100 training trials and 100 test trials each. We refer to the scaled versions of this dataset as \maze-\texttt{L}, \maze-\texttt{M}, \maze-\texttt{S}, respectively. Each scaled dataset contains only 27 conditions and neuron counts ranging from 142 to 162. A number of other Maze datasets are available on DANDI~\cite{maze_datarelease}.

\paragraph{\rtt\hspace{-1mm}.\hspace{-2mm}} Though the Maze datasets contain rich behavior, their stereotypy may still limit the potential complexity of the observed neural signals~\cite{gao2017theory}. Further, natural movements are rarely stereotyped or neatly divided into trials. The random target task dataset~\cite{o2017nonhuman} also contains motor cortical data, but introduces different modeling challenges. It contains continuous, point-to-point reaches that start and end in a variety of locations, do not include delay periods, and have highly variable lengths, with few (if any) repetitions. These attributes preclude simple trial-averaging approaches to de-noise observed spiking activity. Further, we evaluate models using random snippets of the continuous data stream. The unpredictability of random snippets (\eg, a new target could appear at any point in a data window) means that the simplification of autonomous dynamics is a poor approximation (discussed in ~\cite{Pandarinath2018-LFADS,keshtkaran2021large}).

\rtt spans 15 minutes of continuous reaching, artificially split into 1351 600ms ``trials'', and includes 130 neurons and simultaneously monitored hand kinematics. A number of other RTT datasets are available on Zenodo~\cite{o2017nonhuman}. Successful modeling under this benchmark is predicated upon the ability to infer latent representations from single-trial data and infer the presence of unpredictable inputs to the population's activity.

\paragraph{\areatwo\hspace{-1mm}.\hspace{-2mm}} Somatosensory areas may have substantially different dynamics from motor areas~\cite{miri2017behaviorally}, owing to their distinct role of processing sensory feedback, which is critical to our ability to make coordinated movements. \areatwo includes neural recordings from area 2 of somatosensory cortex~\cite{chowdhury2020area}, an area that receives and processes proprioceptive information, or information about the movement of the body. These data were recorded as a monkey performed a simple visually-guided reaching task, where each trial consisted of a reach to a visually presented target using a manipulandum. However, in a random 50\% of trials, the manipulandum unexpectedly bumped the monkey’s arm in a random direction before the reach cue, necessitating a corrective response~\citep{chowdhury2020area}.

\areatwo includes 462 total trials and 65 neurons, and associated hand kinematics and perturbation information. Successful models would likely need to infer inputs to help describe activity after sensory perturbations, and should also be robust to low neuron counts.

\paragraph{\dmfc\hspace{-1mm}.\hspace{-2mm}} Dorso-medial frontal cortex (DMFC) is a high-level cognitive region, which poses unique challenges for LVMs. Neurons in high-level areas show mixed selectivity to different sensory stimuli and movement parameters~\cite{rigotti2013importance}. Furthermore, behavior and task variables in cognitive tasks are usually not directly observable, which makes the task of inferring internal states even more critical. Additionally, the input and output layers of DMFC are not as clearly delineated as the other sensory or motor cortices~\cite{godlove2014microcircuitry}. \dmfc contains recordings from DMFC while a monkey performed a ``Ready-Set-Go'' time-interval reproduction task \cite{sohn2019bayesian}. In the task, animals estimated a time interval between two visual cues and then generated a matching interval by moving their eyes or hands toward the target cue. Uniquely, that time interval is itself a variable that animals had to infer based on sensory information, and then reproduce based on their internal time estimates. The multiple response modalities, target locations, timing intervals, and timing priors resulted in a total of 40 task conditions~\cite{sohn2019bayesian}. 

\dmfc includes 1289 trials and 54 neurons, and associated task timing, condition, and reaction time information. 
High-performing LVMs would likely need to balance input- and internally-driven dynamics, and again be robust to low neuron counts.

\section{NLB '21: Pipeline and Metrics}
In addition to the diverse neural datasets that we provide as part of this work, we introduce an evaluation framework for evaluating LVMs across a number of axes, which may help the community assess how different design choices affect a model's relevance to the variety of potential challenges in neuroscientific applications, and help users from various areas of systems neuroscience determine which approaches are most relevant to their application.

As motivated in the introduction, LVMs characterize coordination in a neural population's activity across space (\ie, neurons) and time, such that high-performing models should be able to predict activity of neuron-time points that they do not have direct access to. Our primary evaluation metric thus measures an LVM's ability to predict the activity of held-out neurons (co-smoothing, described in Section~\ref{cosmooth}). We also use a secondary metric that measures an LVM's ability to predict held-out timepoints (forward prediction, described in Section~\ref{secondary}), and additional secondary metrics that are dataset-specific (behavioral prediction and PSTH-matching, described in Section~\ref{secondary}).

\begin{figure}[t]
    \centering
    \includegraphics[width=1.0\textwidth]{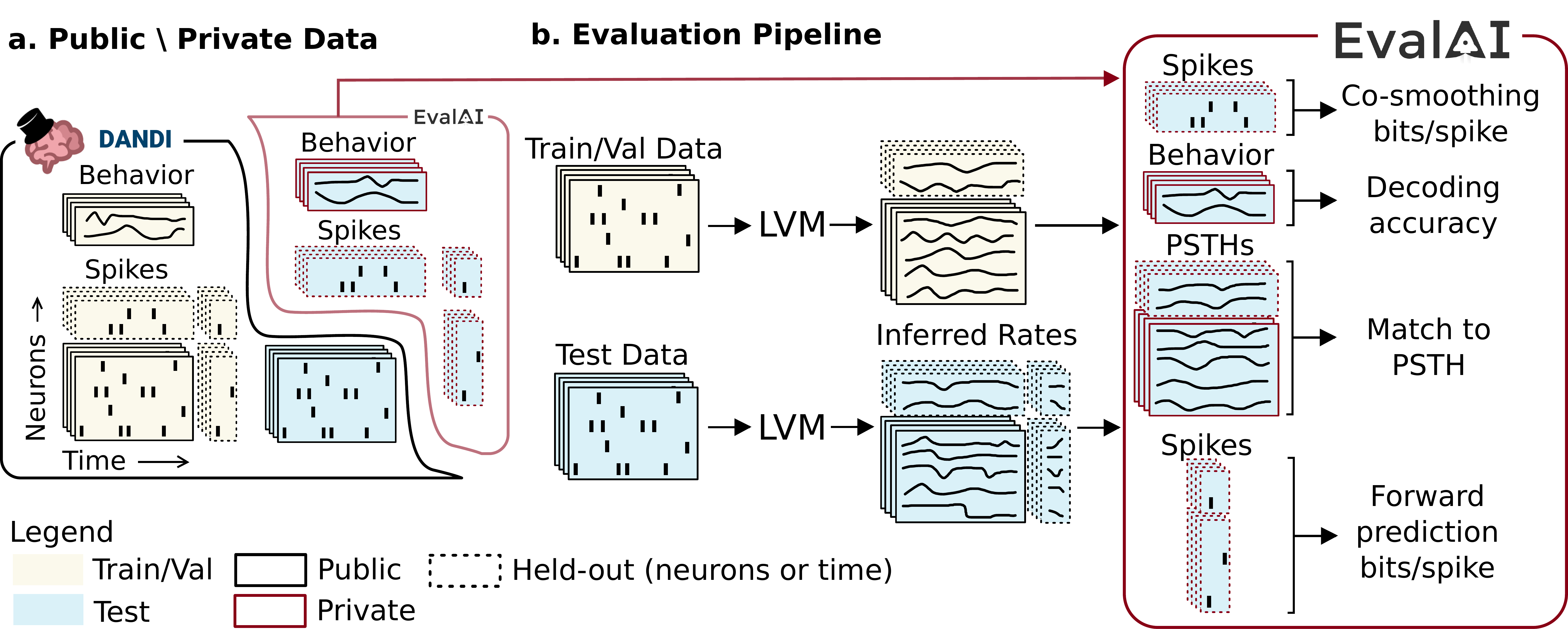}

    \caption{ \footnotesize \xhdr{Evaluation pipeline.} \textbf{a}. Datasets are split into several components to enable rigorous evaluation of the candidate model (described in Section~\ref{sec:eval}). \textbf{b}. During evaluation, users submit firing rate predictions that are evaluated against private data (held on EvalAI servers) via co-smoothing (primary metric, Section~\ref{cosmooth}), as well as match to PSTH, forward prediction, and behavioral decoding (secondary metrics, Section~\ref{secondary}).
    \vspace{-1mm}
    }
    \label{fig:evalai}
\end{figure}

\vspace{-1mm}
\subsection{Evaluation strategy and pipeline}\label{sec:eval}
\vspace{-1mm}

To support evaluation, in addition to defining train/val/test splits for our datasets, we further designate neurons and timepoints as ``held-in'' or ``held-out.'' While both held-in and held-out data, as well as behavioral data, are available for train/val trials, only held-in data are provided for test trials. Withheld test data are stored exclusively on EvalAI servers (Figure~\ref{fig:evalai}).

At evaluation time, users submit inferred firing rates for both train/val and test splits. Train/val inferred rates are only used to compute a simple linear decoder for behavioral prediction - held-out (future) timepoints are unneeded. Test inferred rates are evaluated via co-smoothing (primary metric), as well as match to PSTH, forward prediction, and behavioral decoding using the linear decoder (secondary metrics). Model predictions are submitted on EvalAI in 5 ms bins.

By requiring only that users submit predictions for the specific neuron-time firing rates, we do not place specific restrictions on the method by which candidate LVMs are trained or perform inference. For example, methods may leverage the provided behavioral data as additional information to guide learning of their latent representations, as done in~\citep{Sani2021-ve}. Note also that we specify a val split as a standard for reporting ablations and analyses, though users are free to validate their model with other splits.

\vspace{-1mm}
\subsection{Primary evaluation metric: Co-smoothing}\label{cosmooth}
\vspace{-1mm}

Our primary evaluation metric is the log-likelihood of held-out neurons' activity. 
As mentioned previously, the test data is split into held-in and held-out neurons (Figure~\ref{fig:evalai}a). Given the training data and the held-in neurons in the test data, the user submits the predicted firing rates $\bm{\lambda}$ of the held-out neural activity $\hat{\bm{y}}$. We use a Poisson observation model in the log-likelihood such that the probability of the held-out spike count of neuron $n$ at time point $t$ is $p(\hat{\bm{y}}_{n,t}) = \operatorname{Poisson}(\hat{\bm{y}}_{n,t}; \bm{\lambda}_{n,t}).$ The overall log-likelihood $\mathcal{L}(\bm{\lambda}; \hat{\bm{y}})$ is the sum of the log-likelihoods over all held-out neurons $n$ and time points $t$. 

To normalize the log-likelihood score, we convert it to bits per spike using the mean firing rates of each held-out neuron \cite{pillow2008spatio}. Bits per spike is computed as follows:
\[
    \text{bits}/\text{spike} = \frac{1}{n_{sp} \log 2}(\mathcal{L}(\bm{\lambda};\hat{\bm{y}}_{n,t}) - \mathcal{L}(\bar{\bm{\lambda}}_{n,:};\hat{\bm{y}}_{n,t}))
\]
where $\bar{\bm{\lambda}}_{n,:}$ is the mean firing rate for the neuron $n$ and $n_{sp}$ is its total number of spikes. 
Thus, a positive bits per spike (bps) value indicates that the model infers a neuron’s time-varying activity better than a flat mean firing rate. 

The approach of predicting the activity of held-out neurons conditioned on the held-in neurons on test data is referred to as co-smoothing \cite{Macke2011-ep} and provides a generalization of leave-neuron-out accuracy measures now commonly used in the neuroscience community \cite{Macke2011-ep,Yu2009-qp,Pandarinath2018-LFADS,Zhao2016a,Wu2017-az}. Co-smoothing assumes that the latent variables underlying the activity of held-out neurons can be inferred from the training neurons supplied at test time. While this may not hold for all neurons in a population, the observation that latent representations are distributed across many neurons provides strong support for this assumption (reviewed in~\citep{Yu2009-qp,cunningham2014dimensionality}).

\vspace{-1mm}
\subsection{Secondary metrics}\label{secondary}
\vspace{-1mm}
{\bf Behavioral decoding.} Relating neural activity to behavior is a common goal in modeling neural data, and neural latent variables are often interpreted by identifying relationships between the latent states and behavioral variables of interest. Therefore, we included behavioral decoding accuracy given the inferred latent variables as a secondary benchmark metric. 

For the \maze, \rtt, and \areatwo datasets, behavioral decoding is evaluated by fitting a ridge regression model from training rates to behavioral data and measuring R2 of predictions from test rates. Though better decoding performance can be achieved with more complex decoders, we choose to enforce a linear mapping for all models to prevent excessively complex decoders from compensating for poor latent variable estimation. With all three datasets, the behavioral data used is monkey hand velocity; many kinematic variables are known to correlate with sensorimotor activity, but some (such as position) change slowly, making it easy to saturate decoding performance by heavy smoothing of the neural activity. We hypothesized that hand velocity would provide a challenging enough decoding target to differentiate the quality of the representations inferred by different models.  

For \dmfc, behavioral decoding is more difficult to evaluate (as described earlier). However, previous work has indicated that the rate at which the neural population state changes, or the neural speed, correlates negatively with $t_{p}$, the time between the Set cue and the monkey's Go response in the Ready-Set-Go (RSG) task \cite{sohn2019bayesian,keshtkaran2021large}.
Thus, we compute average neural speed for each trial from test rate predictions, and then calculate Pearson's \textit{r} between neural speeds and the measured $t_{p}$, as a measure of how well predicted neural activity reflects observed behavior~\citep{sohn2019bayesian}.

{\bf Match to PSTH.} When behavioral tasks include distinct conditions with repeated trials, a coarse, commonly-used method to de-noise spiking activity is to compute a peri-stimulus time histogram (PSTH). PSTHs are computed by averaging neuronal responses across trials within a given condition, and thus reveal features that are consistent across repeated observations of the behavioral condition~\cite{cunningham2014dimensionality}. For LVMs, inferring firing rates that recapitulate the PSTHs provides evidence that a model can capture certain stereotyped features of the neurons' responses. However, we use the match to PSTH as a secondary metric because of two limitations: First, not all datasets are well-suited to computing PSTHs. And second, the PSTH treats all across-trial variability as ``noise'', and thus matching the PSTH does not test a model's ability to predict single-trial variability that may be prominent and a key part of a given brain area's computational role~\cite{cunningham2014dimensionality}. 

Outside of the \rtt dataset, which is not well-suited to trial-averaging approaches due to its lack of clear trial structure, typical neural responses to specific conditions for the other datasets can be estimated by averaging smoothed spikes across trials within the same condition.  We evaluate how well predicted rates match PSTHs by computing $R^2$ between trial-averaged model rate predictions for each condition and the true PSTHs. $R^2$ is first computed for each neuron across all conditions and then averaged across neurons.

{\bf Forward prediction.} We additionally test the model's ability to predict the responses of \textit{all neurons} at unseen, future time points. The forward prediction benchmark is evaluated in a similar manner as co-smoothing, with the distinction that the held-out responses are across all neurons at future time points. 
Forward prediction provides a further measure of how well a model can capture temporal structure in the data. However, it assumes that future neural activity can be predicted based on prior neural activity. This should not be the case in general, and is especially problematic for brain regions and behavioral tasks in which unpredictable inputs are common. Thus while forward-prediction provides some assessment of an LVM's ability to predict activity that is itself predictable, it is best applied to scenarios where data can be well-modeled via autonomous dynamics (such as \maze\hspace{-1mm}).

\vspace{-1mm}
\subsection{Baselines to seed the benchmark}
\vspace{-1mm}
We seed the benchmark with 5 established methods for modeling neural population activity: Smoothed spikes \cite{Yu2009-qp}, Gaussian Process Factor Analysis (GPFA) \cite{Yu2009-qp}, Switching Linear Dynamical System (SLDS) \cite{Linderman2017-RSLDS,Nassar2018b,Zoltowski2020-up}, AutoLFADS \cite{keshtkaran2021large}, and the Neural Data Transformer (NDT) \cite{ye2021representation}.

These models vary in terms of their underlying assumptions on dynamics, the forward (or generative) model, and their overall complexity \cite{hurwitz2021building}. Spike smoothing is the simplest yet still common approach for de-noising firing rates by convolving spiking activity with a Gaussian kernel. GPFA models neural state as a low-dimensional collection of Gaussian processes and thus imposes a simple linear dynamics on the latent space of the generative model. SLDS expands upon the standard linear dynamical system and approximates complex non-linear dynamics by alternating or ``switching'' between multiple distinct linear systems. AutoLFADS is a variational sequential autoencoder that models neural dynamics with an RNN and thus can capture complex nonlinear dynamics and embeddings. NDT uses a transformer to generate neural activities, without any explicit dynamics model.

\vspace{-1mm}
\section{Results}
\vspace{-1mm}
\subsection{\maze as Neuroscience MNIST}

\begin{figure}[t]
    \centering
    \includegraphics[width=0.98\textwidth]{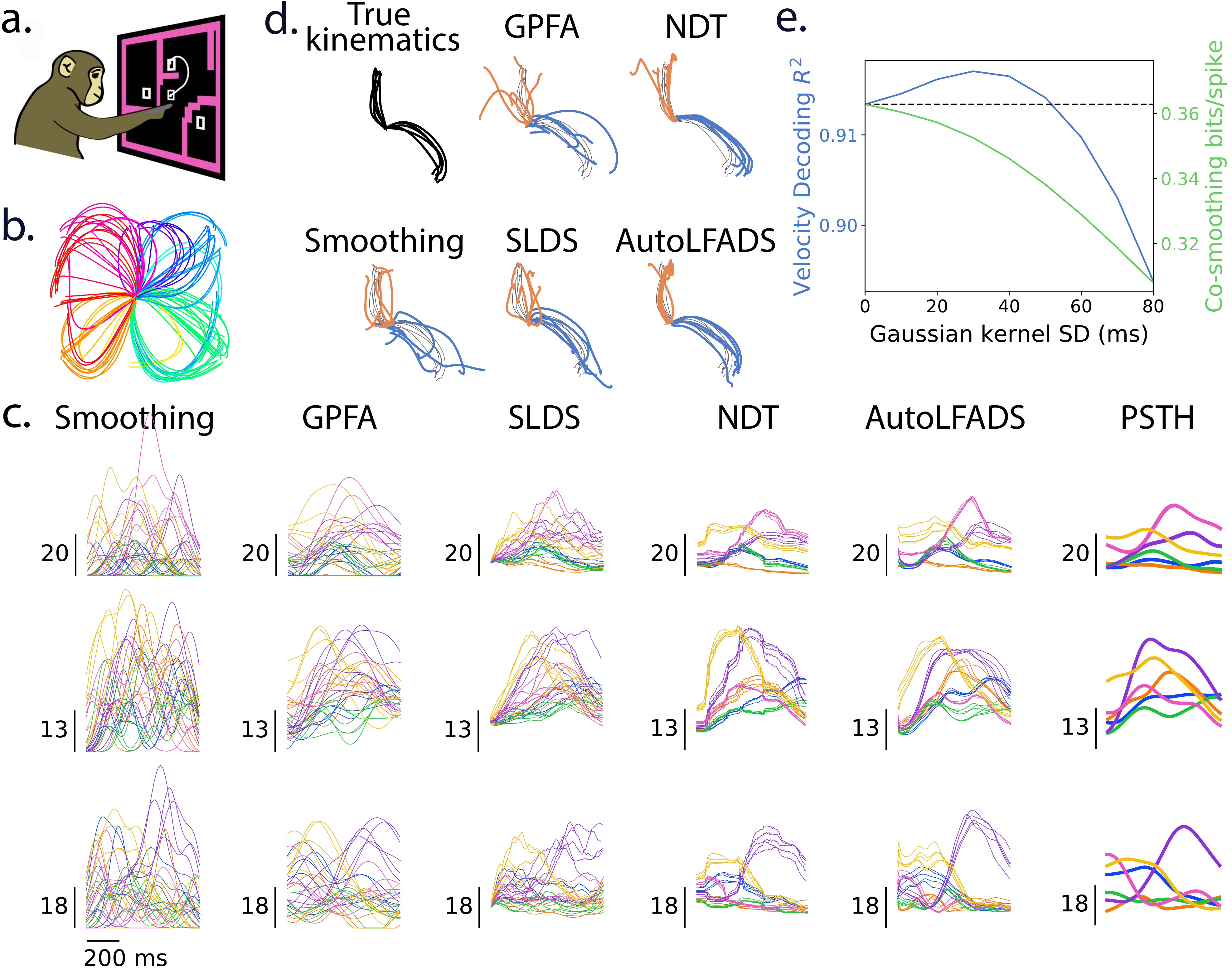}

    \caption{\footnotesize\xhdr{\maze in detail}. \textbf{a}. A monkey made curved reaches through a virtual maze. \textbf{b}. The task contained 108 different reaching conditions (colored by angle to target). \textbf{c}. (columns 1-5) Single trial inferred rates from the baseline models, and (column 6) PSTHs, for 3 neurons and 6 conditions (out of 108). Vertical scale shows spikes/second. Higher-performing models infer single-trial rates with greater across-trial consistency, that also resemble the trial-average. \textbf{d}. Single-trial hand trajectories are estimated via linear decoding of hand velocities based on inferred rates, and overlaid on the measured hand trajectories (grey). Shown are 5 trials each for 2 conditions. \textbf{e}. Behavioral decoding performance can be improved by smoothing the rates inferred by AutoLFADS, though co-smoothing performance decreases.
    }
    \label{fig:maze_workup}
\end{figure}

We first provide a close examination of model results on \maze, which has many desirable qualities for a basic model litmus test. It has rich structure with 108 different reaching conditions, 180 neurons, and $>2000$ trials. Yet each condition also has many repeated trials, allowing assessment through PSTH metrics. Additionally, because motor cortex has close anatomical ties to motor output, \maze supports model evaluation via behavioral decoding. Finally, it is well-established both empirically and theoretically, as many LVMs have already been evaluated on the dataset and the dynamics are well-approximated as autonomous~\cite{Pandarinath2018-LFADS}, inviting forward prediction measures to assess the quality of an LVM's dynamics model. This well-controlled setting makes \maze like a ``Neuroscience MNIST'': immediately useful for validation of neural LVMs, even if potentially limited for long term neuroscientific or modeling advances. The evaluation of our baseline models on this dataset is presented in~\figref{fig:maze_workup} and~\tableref{tab:maze}.

\begin{table}[t]
    \centering
    \resizebox{0.85\linewidth}{!}{
        \begin{tabular}{@{}lccccc@{}}
            & Co-smoothing bps $(\uparrow)$ & Behavior decoding $(\uparrow)$ &
            PSTH $R^2$ $(\uparrow)$ &
            Forward pred bps  $(\uparrow)$&
            \\[0.05in]
            \toprule
Smoothing &
$0.211$  & 
$0.624$ & 
$0.183$ & 
$-$  & 
\\
GPFA &
$0.187 $\scriptsize{$\pm 0.001$} & 
$0.640$\scriptsize{$\pm 0.001$} & 
$0.518$\scriptsize{$\pm 0.002$} & 
$-$ & 
\\
SLDS &
$0.219 $\scriptsize{$\pm 0.006$} &
$0.775$\scriptsize{$\pm 0.006$} & 
$0.482$\scriptsize{$\pm 0.036$} & 
$-1.020$\scriptsize{$\pm 0.943$} & 
\\
NDT &
$0.329$\scriptsize{$\pm 0.005$} & 
$0.897$\scriptsize{$\pm 0.009$} & 
$0.579$\scriptsize{$\pm 0.023$} & 
$0.209$\scriptsize{$\pm 0.010$} & 
\\
AutoLFADS &
$0.346$\scriptsize{$\pm 0.005$} & 
$0.907$\scriptsize{$\pm 0.002$} & 
$0.631$\scriptsize{$\pm 0.022$} & 
$0.239$\scriptsize{$\pm 0.003$} & 
\\
        \bottomrule
    \end{tabular}
    }
    \vspace{5pt}
    \caption{\footnotesize \xhdr{\maze metrics}. 
    All NLB'21 metrics $\pm$ standard error of the mean applied to \maze\hspace{-1mm}. Baseline rank order is consistent across metrics; deep neural networks (NDT and AutoLFADS) improve on less expressive baselines. High decoding performance across the board directly links between recorded activity and external behavior.
    }
    \vspace{-15pt}
    \label{tab:maze}
\end{table}

For the baseline models, the rank ordering of performance is largely consistent across metrics, and deep networks (AutoLFADS, NDT) are stronger across the board. The consistency of rankings in \maze demonstrates the validity of co-smoothing in a well-structured dataset, motivating its use in more challenging ones. 

Note that we do not expect primary and secondary metrics to be perfectly correlated. For example, post-processing the AutoLFADS-inferred rates via smoothing decreases co-smoothing performance, but increases behavioral decoding performance (~\figref{fig:maze_workup}e). This highlights a conflict between unsupervised (co-smoothing) and supervised (behavioral decoding) evaluations. Similarly, we observe that GPFA underperforms spike smoothing when assessed via co-smoothing, but outperforms it in behavioral decoding; following~\figref{fig:maze_workup}e, this may be because GPFA has inferred rates with coarse features that evolve even slower than the spike smoothing kernel, which is consistent with the single-trial inferences in~\figref{fig:maze_workup}c. %
Note also that further tuning of hyperparameters or the kernel parameterization could improve GPFA, and more advanced versions that are designed for spike count data \cite{Zhao2017c} may achieve higher performance.

We also separately examine forward prediction performance; they are omitted for spike smoothing and GPFA which do not have clear adaptations for forward dynamics. Absolute performance of likelihood is generally uninterpretable, but SLDS scores are negative, implying poor fitting that underperforms a simple estimate that uses the average firing rates. We expect that the recurrent version of SLDS, an rSLDS, would improve on forward prediction performance. NDT and AutoLFADS both perform well beyond this null hypothesis, and comparing their forward prediction relative to co-smoothing, we conclude they are inferring the predictable, autonomous dynamics we expect in \maze.

\vspace{-1mm}
\subsection{Results across datasets}

\begin{table}[t]
    \centering
    \resizebox{1.0\linewidth}{!}{
        \begin{tabular}{@{}lccccccc@{}}
            & \maze & 
            \mazeL & \mazeM & \mazeS &
            \rtt &
            \areatwo & \dmfc
            \\[0.05in]
            \toprule
Smoothing &
$0.211$  & 
$0.224$ & $0.167$ & $0.191$ &
$0.147$ & 
$0.154$ & 
$0.120$
\\
GPFA &
$0.187$\scriptsize{$\pm 0.001$}  & 
$0.239$\scriptsize{$\pm 0.001$} & $0.172$\scriptsize{$\pm 0.001$} & $0.217$\scriptsize{$\pm 0.002$} &
$0.155$\scriptsize{$\pm 0.000$} & 
$0.168$\scriptsize{$\pm 0.000$} & 
$0.118$\scriptsize{$\pm 0.000$}
\\
SLDS &
$0.219$\scriptsize{$\pm 0.006$}  & 
$0.290$\scriptsize{$\pm 0.008$} & $0.210$\scriptsize{$\pm 0.008$} & $0.250$\scriptsize{$\pm 0.001$} &
$0.165$\scriptsize{$\pm 0.004$} & 
$0.187$\scriptsize{$\pm 0.005$} & 
$0.120$\scriptsize{$\pm 0.001$}
\\
NDT &   
$0.329$\scriptsize{$\pm 0.005$}  & 
$0.362$\scriptsize{$\pm 0.006$} & $0.259$\scriptsize{$\pm 0.021$} & $0.251$\scriptsize{$\pm 0.014$} &
$0.160$\scriptsize{$\pm 0.009$} & 
$0.267$\scriptsize{$\pm 0.003$} & 
$0.162$\scriptsize{$\pm 0.009$}
\\
AutoLFADS &
$0.346$\scriptsize{$\pm 0.005$}  & 
$0.374$\scriptsize{$\pm 0.005$} & $0.304$\scriptsize{$\pm 0.006$} & $0.301$\scriptsize{$\pm 0.009$} &
$0.192$\scriptsize{$\pm 0.003$} & 
$0.259$\scriptsize{$\pm 0.001$} & 
$0.181$\scriptsize{$\pm 0.001$}
\\
            \bottomrule
        \end{tabular}
    }
    \vspace{5pt}
    \caption{\footnotesize\xhdr{Co-smoothing across datasets}. We report co-smoothing bits/spike $\pm$ standard error of the mean across the datasets in NLB. While exact rankings and performance gaps vary per dataset, more expressive deep neural networks tend to perform the best. Note that absolute co-smoothing performance is not easily compared across datasets. \vspace{-3mm}
    }
    \label{tab:cosmoothing}
\end{table}

\vspace{-1mm}
The generality of co-smoothing allows its application across the diverse data from different brain regions (\tableref{tab:cosmoothing}). The new datasets preserve overall rankings from \maze, but provide more nuance. In contrast to the \maze and \areatwo, which both show a clear gap in results between linear and deep models, we observe a smaller gap in \rtt. \dmfc in particular demonstrates the need for deep, expressive LVMs in identifying structure in deep cognitive areas, since only deep networks identify structure better than the simple spike smoothing baseline.

The different datasets have varied numbers of neurons with variable firing rates. As such, direct comparisons of scores across datasets is tricky. However, by using multiple datasets from the same underlying experiment (as in \texttt{MC\_Maze-\{S,M,L\}}), we can make more direct comparisons and also study the scaling of different methods across different data sizes. We find that both the NDT and AutoLFADS scale well across small sample sizes, with AutoLFADS providing more stable performance on smaller datasets and NDT lagging slightly for \mazeS.

\vspace{-1mm}
\section{Discussion}

\vspace{-1mm}
NLB'21 is a substantial initial step towards standardized evaluation  of LVMs on neural data. We hope that in highlighting the ability of LVMs to extract population structure across the brain, we can catalyze the broad adoption of LVMs in systems neuroscience.

Our model-agnostic framework enables standardized comparison of 5 approaches that vary in complexity and assumptions about latent structure, across 7 different real neuroscientific datasets. This baseline evaluation is already, to our knowledge, the broadest LVM comparison yet attempted, and the platform will allow the comparison to be greatly extended via community-driven contributions. 

Our initial evaluation delineates deep networks as powerful tools for uncovering structure from neural population activity. We anticipate the benchmark as a conduit for ML researchers with cutting-edge approaches to evaluate those approaches on neuroscientific datasets. To be inclusive to these methods, we do not preclude alternate model training strategies (\eg data augmentation, transfer learning, etc.), nor do we restrict by computational or memory requirements, though we do encourage reporting these details on submission.

\xhdr{Extensions and Limitations.}
NLB'21 is our first suite in a planned effort to rigorously evaluate models of neural data. We anticipate extensions, for example, towards modeling multi-region interactions, alternate recording modalities, transfer learning, and robust state estimation under recording instability. A particularly important direction for future work, and a subject of interest for the broader ML community, is developing measures for evaluating model or latent state interpretability, as interpretability is key to an LVM's ability to drive scientific progress.

\xhdr{Ethical considerations.}
All datasets were collected under procedures and experiments that were approved by the Institutional Animal Care and Use Committees at the respective institutions. Specifics of the experimental procedures can be found in the primary references for each dataset. Animal models are a cornerstone of research to increase our understanding of the brain, and we hope that benchmarking efforts that improve standardization, such as NLB, might help minimize redundancy in data collection, while also enabling the development of LVMs that allow researchers to get more use from existing data.

Separately, advances in LVMs might be used to improve the performance of brain-machine interfaces or other therapeutic devices related to brain injury or disease~\citep{pandarinath2018latent}. While such devices are largely targeted towards restoring function to people with disabilities or impairments, it is important to consider the potential impacts on fundamental aspects of the human experience, such as identity, normality, authority, responsibility, privacy, and justice, which are the subject of ongoing study ~\citep{klein2015engineering,pandarinath2021science}.

\bibliography{neurips}

\iftoggle{neurips}{
    
\newpage

\appendix

\section{Appendix}

\subsection{Additional Analyses}

\subsubsection{Maze co-smoothing and decoding}

\begin{figure}[htbp]
\centering
\includegraphics[width=0.5\textwidth]{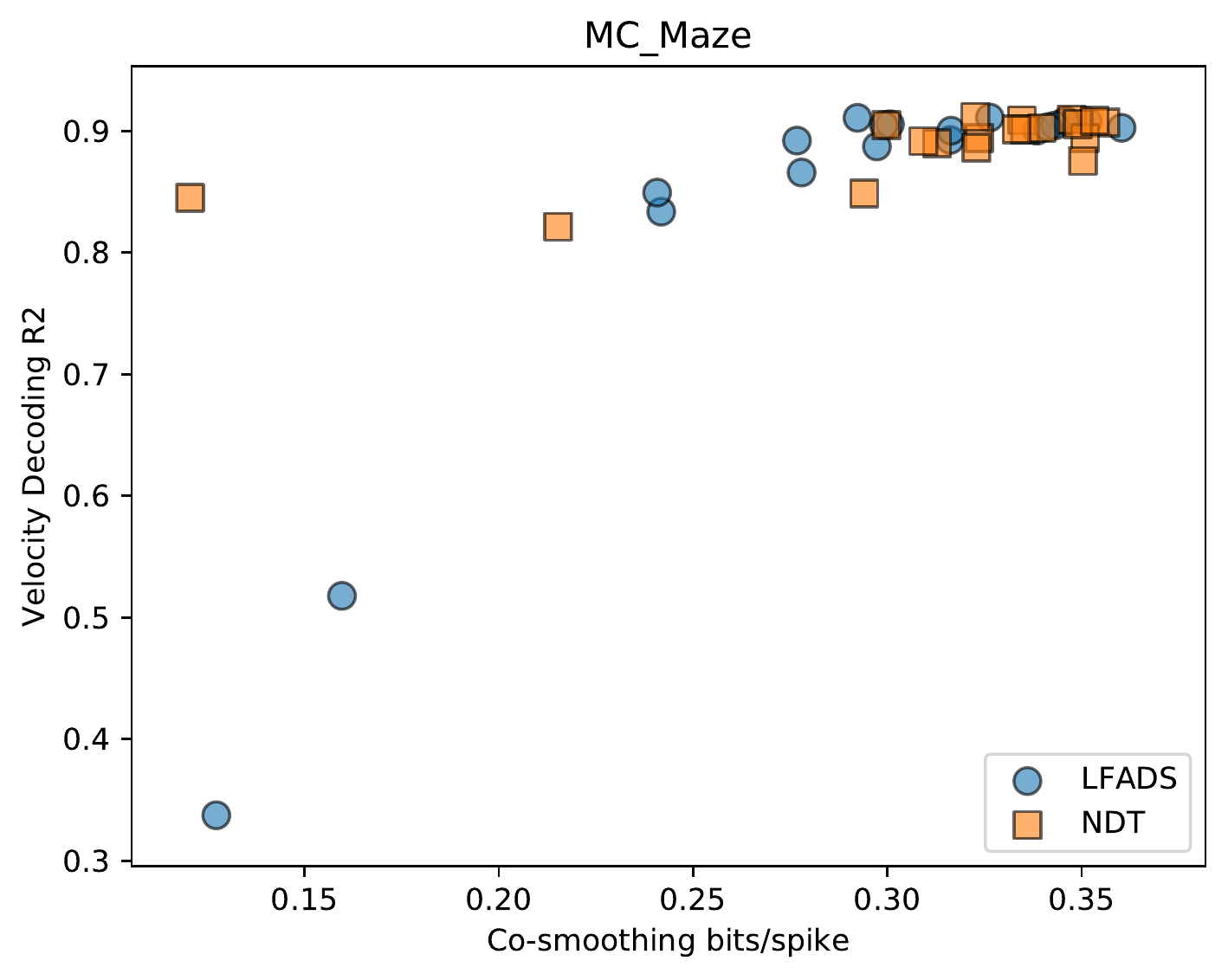}
\caption{\footnotesize \xhdr{Performance on co-smoothing and velocity decoding in \maze}. Random search LFADS and NDT models show good behavior decoding is possible under a wide range of fits to neural data, as quantified by co-smoothing performance.}
\label{fig:mazerandsearch}
\end{figure}

In order to examine the relationship between co-smoothing and behavioral decoding, we trained 20 LFADS and 20 NDT models with random parameters on the \maze dataset and evaluated their performance. As seen in Figure~\ref{fig:mazerandsearch}, both methods easily reach an apparent upper bound in decoding accuracy at around $0.9$ but display a wide range of co-smoothing scores in that region. This demonstrates that there is an additional dimension to the models that is captured by co-smoothing but not by behavioral decoding.

\subsubsection{Co-smoothing bootstrap analysis}

To evaluate the reliability of the co-smoothing metric, we use a bootstrap analysis where we compare model performances pairwise on 100 bootstrap samples of the test set.

\newpage

\setlength{\textfloatsep}{0.0pt}
\begin{figure}[t!]
    \centering
    \includegraphics[width=0.92\textwidth]{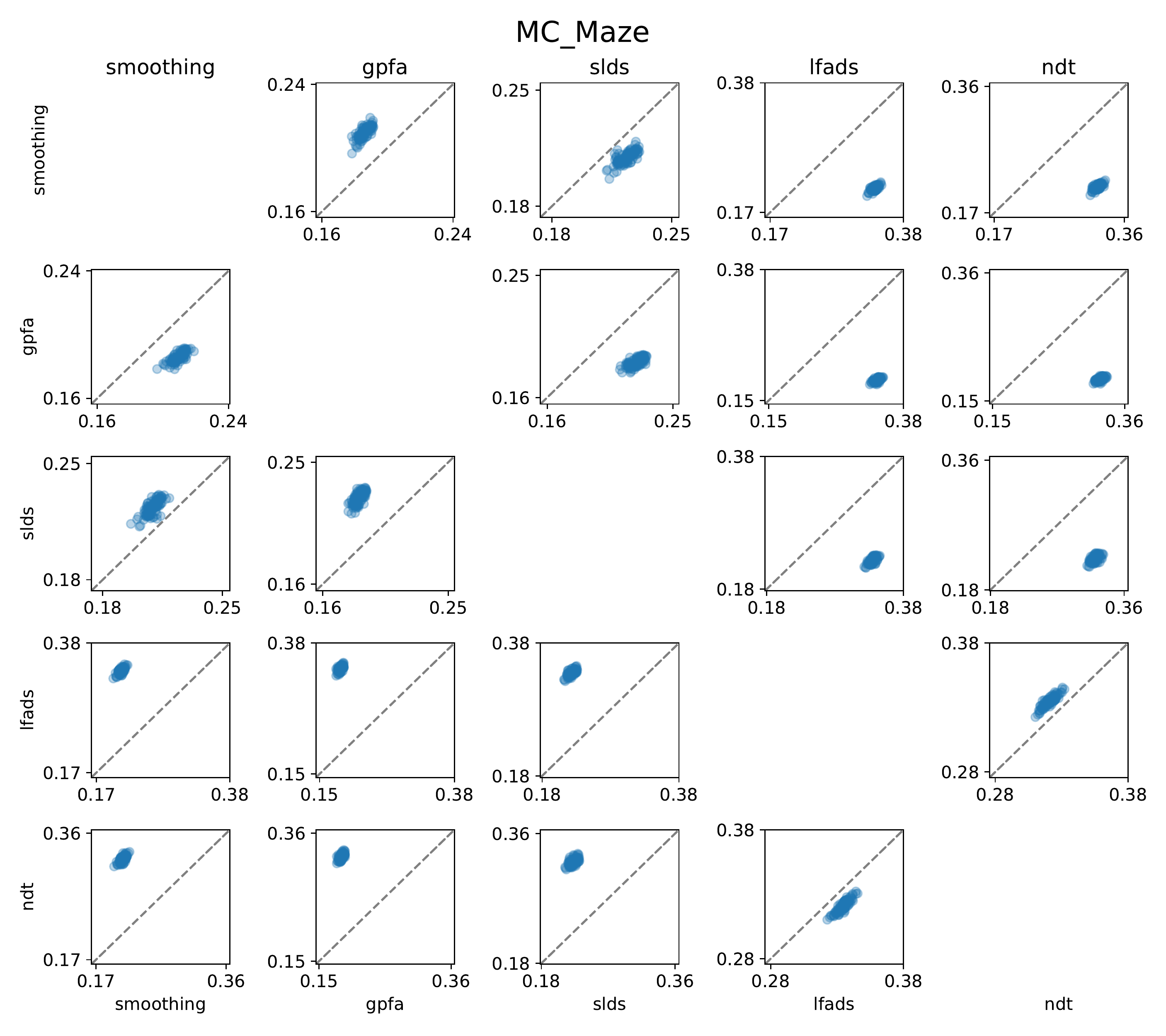}
\end{figure}
\begin{figure}[b!]
    \centering
    \includegraphics[width=0.92\textwidth]{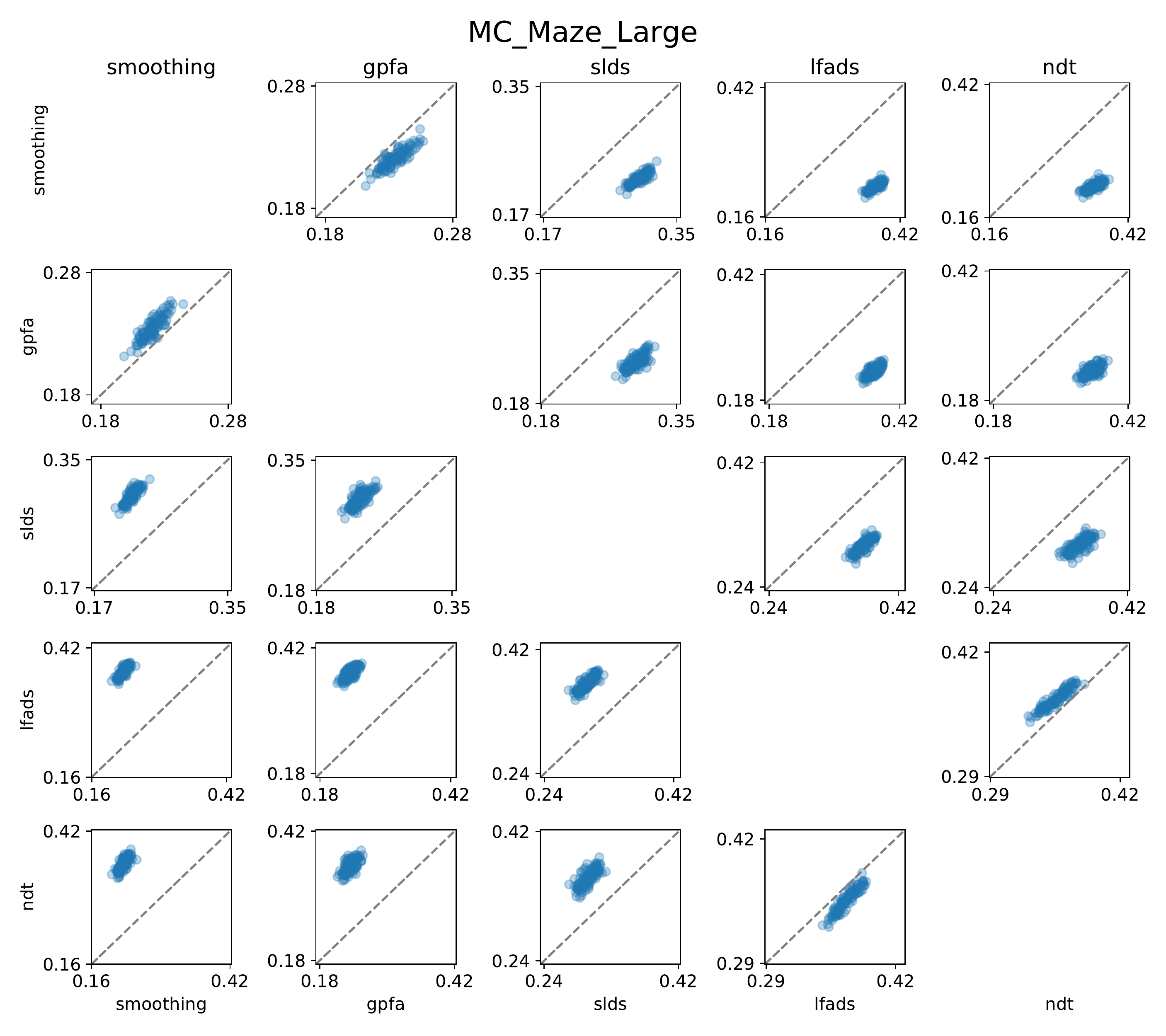}

\end{figure}
\newpage
\begin{figure}[t!]
    \centering
    \includegraphics[width=0.92\textwidth]{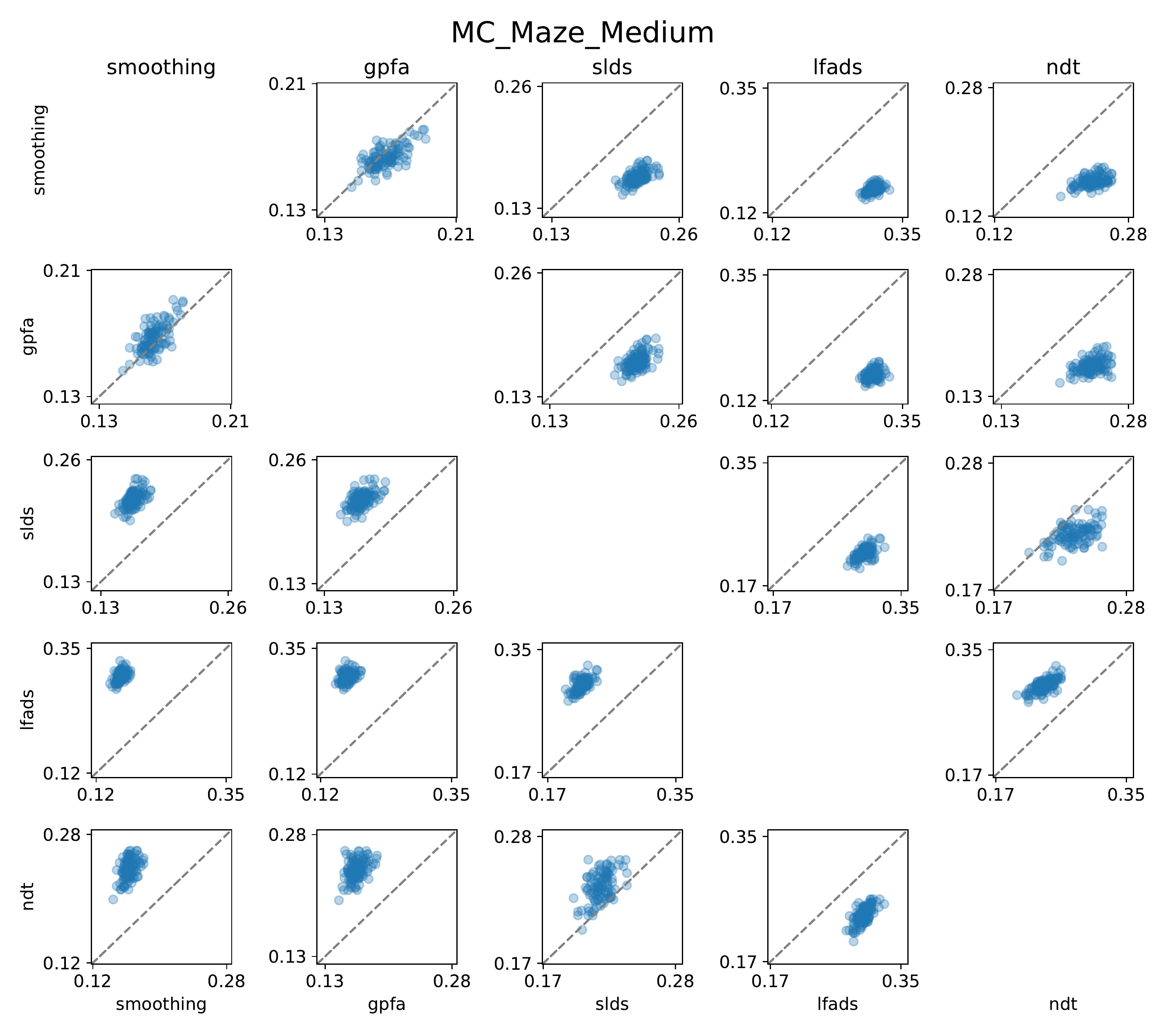}

\end{figure}
\begin{figure}[b!]
    \centering
    \includegraphics[width=0.92\textwidth]{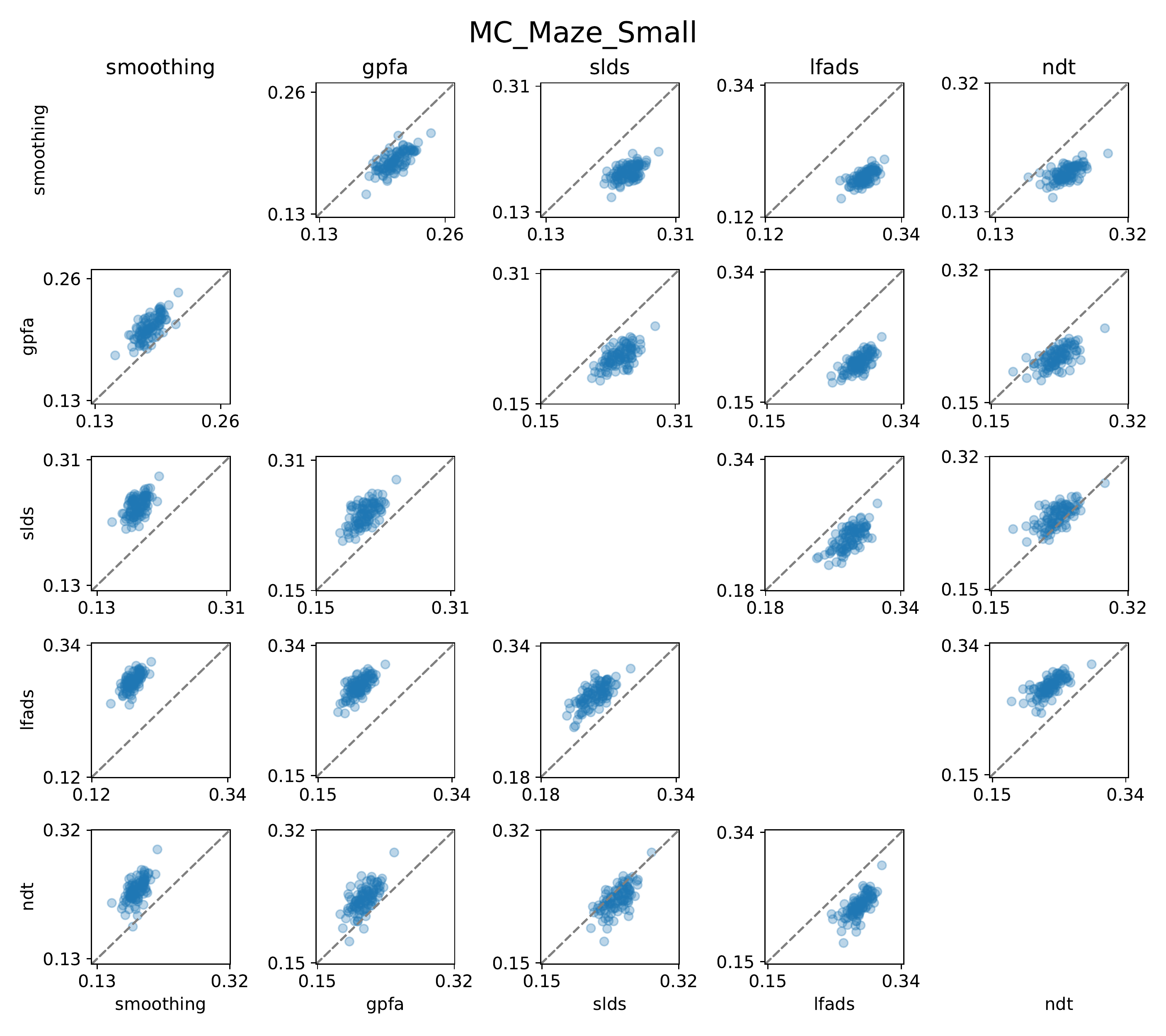}

\end{figure}
\newpage
\begin{figure}[t!]
    \centering
    \includegraphics[width=0.92\textwidth]{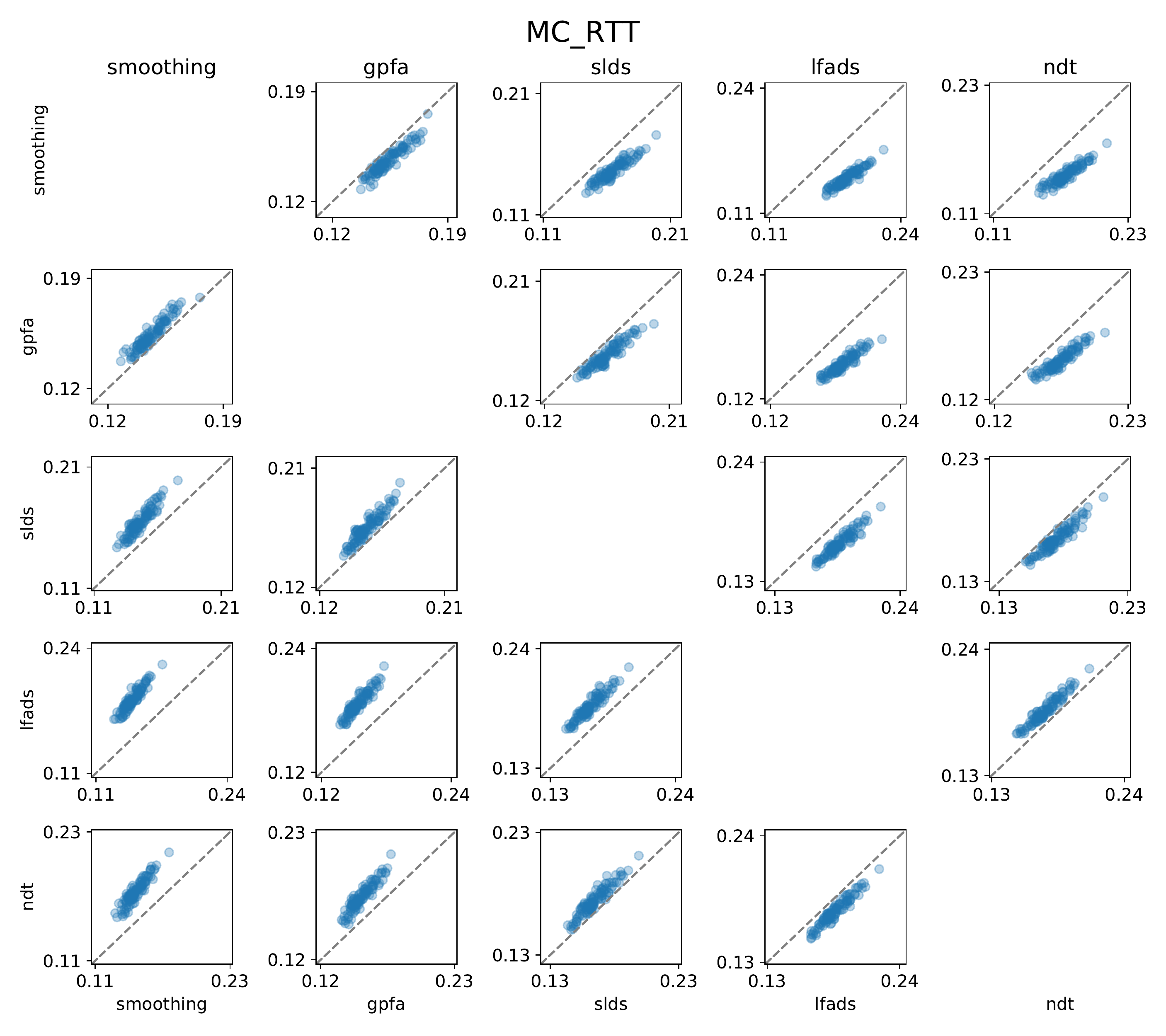}

\end{figure}
\begin{figure}[b!]
    \centering
    \includegraphics[width=0.92\textwidth]{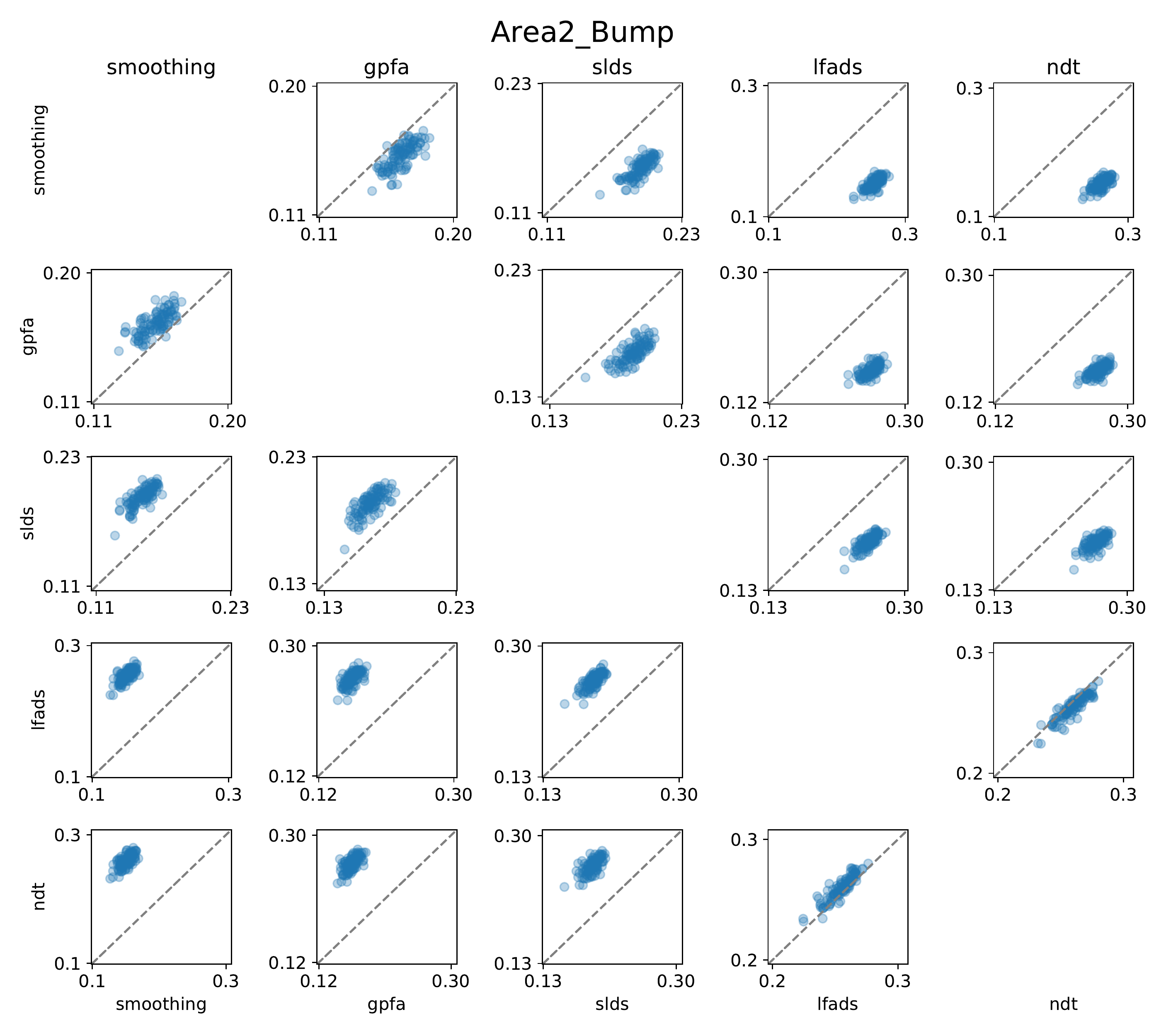}

\end{figure}
\newpage
\setlength{\textfloatsep}{20.0pt plus 2.0pt minus 4.0pt}
\begin{figure}[t!]
    \centering
    \includegraphics[width=0.92\textwidth]{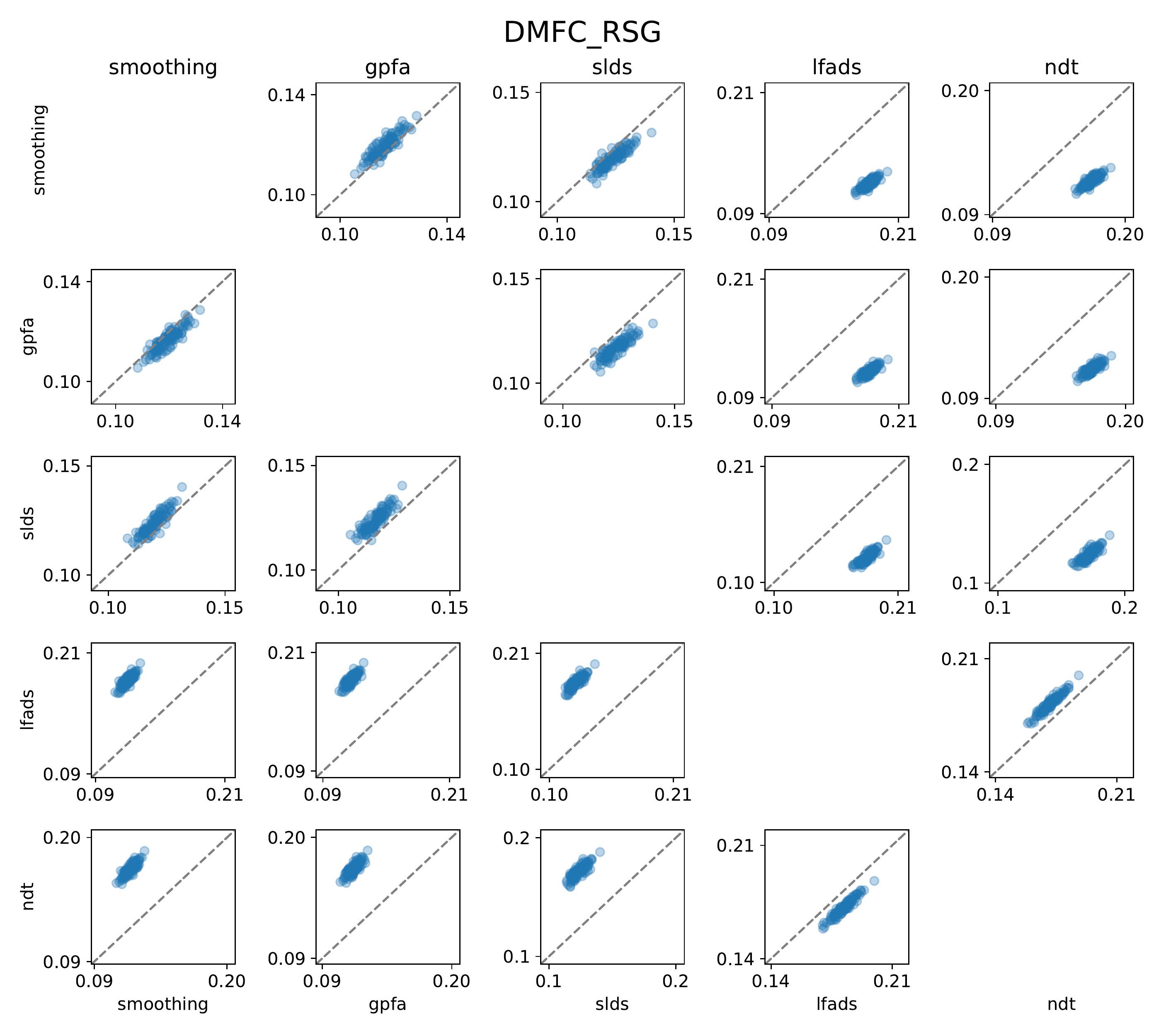}
    
\caption{\footnotesize\xhdr{Model co-smoothing performance compared across bootstrap samples.} Each panel compares the performance of two models, and within a panel, each point indicates the performance of the two models on the same bootstrap sample. The fact that most points lie on one half of the diagonal indicates that model rankings are robust to the particular evaluation dataset we used.}
\label{fig:bootstrapscatter}
\end{figure}

\newpage

\subsection{The rapid growth of neural LVMs}

\begin{table}
\resizebox{1.00\textwidth}{!}{
\begin{threeparttable}
    
    \begin{tabular}{rlllccccc}
    \toprule
	\hiderowcolors
	& & & & \multicolumn{5}{c}{Metrics} \\
	\multirow[c]{2}{*}{Year}
	    &
	\multirow[c]{2}{*}{Venue}
	    &   
	\multirow[c]{2}{*}{Work}
	    & 
    \multirow[c]{2}{*}{Dataset}
	    & 
    \multirow[c]{2}{*}{\shortstack{Held-out \\ Channel}}
        &
    \multirow[c]{2}{*}{\shortstack{Behavior/ \\ Stimulus}}
	    &
    \multirow[c]{2}{*}{PSTH}
	    & 
    \multirow[c]{2}{*}{\shortstack{Held-out \\ Time}}
	    &
    \multirow[c]{2}{*}{Other}
	\\
	\\
	\midrule
	08 & NeurIPS & Yu \etal \citep{yu_neurips_08} & \citep{churchland_neural_2006} & \cmark &  &  &  &  \\
	11 & NeurIPS & Macke \etal \citep{Macke2011-ep} & \citep{churchland_neural_2006} & \cmark &  &  &  &  \\
	11 & NeurIPS & Petreska \etal \citep{Petreska2011-be} & \citep{churchland_neural_2006} & & \cmark &  &  & \\
	12 & NeurIPS & Buesing \etal \citep{buesing_neurips_12} & \citep{churchland_neural_2006} & \cmark &  &  &  &  \\
	13 & NeurIPS & Turaga \etal \citep{turaga_neurips_13} & \citep{turaga_neurips_13} &  &  &  & \cmark & \cmark \\
	13 & NeurIPS & Pfau \etal \citep{Pfau2013-oi} & \citep{koyama2010approximate} & & & & & \cmark \\
	14 & NeurIPS & Semedo \etal \citep{semedo_neurips_14} & \citep{semedo_neurips_14} & \cmark & & & & \\
	15 & NeurIPS & Gao \etal \citep{gao_neurips_15} & \citep{churchland_neural_2006} & \cmark & & & & \\
	16 & NeurIPS & Gao \etal \citep{gao_neurips_16} & \citep{graf2011decoding,churchland_neural_2006} & & & \cmark & \cmark & \\
	17 & NeurIPS & \makecell[l]{Nonnenmacher \\ \etal \citep{nonnenmacher_neurips_17}} & \citep{ahrens_whole-brain_2013} & & & & & \cmark \\
	17 & NeurIPS & Wu \etal \citep{Wu2017-az} & \citep{karlsson_05} & \cmark & \cmark \\
	18 & NeurIPS & Wu \etal \citep{wu_olfactory_2018} & \citep{wu_olfactory_2018} & & & & & \cmark \\
	18 & ICML & Duncker \etal \citep{duncker_neurips_18} & \citep{churchland_neural_2006} & \cmark & \cmark & & & \\
	19 & ICLR & Nassar \etal \citep{Nassar2018b} & \citep{graf2011decoding} & & \cmark & & & \\
	19 & ICLR & \makecell[l]{Farschian\\ \etal \citep{farschian_18_anma}} & \citep{farschian_18_anma} & & \cmark & & & \\
	19 & UAI & She, Wu \citep{She2020-gr} & \citep{graf2011decoding} & \cmark & & & & \\
	19 & NeurIPS & Lee \etal \citep{lee_neurips_19} & \citep{dyer_cryptography-based_2017} & & \cmark & & & \\
	19 & NeurIPS & \makecell[l]{Loaiza-Ganem \\ \etal \citep{loaiza_2019_neurips}} & \citep{churchland_neural_2006} & \cmark & \cmark & \cmark & & \\
	19 & NeurIPS & \makecell[l]{Keshtkaran, \\ Pandarinath \citep{loaiza_2019_neurips}} & \citep{Pandarinath2018-LFADS} & \cmark & \cmark & \cmark & & \\
	19 & NeurIPS & Schein \etal \citep{schein_neurips_19} & \citep{vyas_neural_2018} & & & & & \cmark \\
	20 & ICML & \makecell[l]{Zoltowski\\ \etal \citep{Zoltowski2020-up}} & \citep{yates_functional_2017} & \cmark & & & & \\
	20 & NeurIPS & Rutten \etal \citep{rutten_neurips_20} & \citep{Churchland2012-JPCA} & \cmark & & & & \\
	20 & NeurIPS & Jensen \etal \citep{jensen_neurips_20} & \citep{turner-evans_daniel_kir.zip_2020} & \cmark & & & & \\
	20 & NeurIPS & Zhou, Wei \citep{zhou_neurips_20} & \citep{gallego_long-term_2020} & & \cmark & \cmark & & \\
	20 & NeurIPS & Glaser \etal \citep{glaser_neurips_20} & \citep{gallego_long-term_2020} & & & & \cmark & \\
	20 & NeurIPS & Keeley \etal \citep{keeley_icml_20} & \citep{graf2011decoding,yu_mesoscale_2019} & & & & & \cmark \\
	20 & ICML & Keeley \etal \citep{keeley_neurips_20} &  \citep{yates_functional_2017,yu_mesoscale_2019} & & & & & \cmark \\
	21 & ICML & Kim \etal \citep{Kim2021icml} & \citep{Kim2021icml} & & \cmark & & &  \\
	21 & NeurIPS & Liu \etal \citep{liu_neurips_21} & \citep{dyer_cryptography-based_2017} & & \cmark & & & \cmark \\
	21 & NeurIPS & Smith \etal \citep{smith_neurips_21} & \citep{Churchland2012-JPCA} & & & \cmark & & \cmark \\
	21 & NeurIPS & Zhu \etal \citep{zhu_neurips_21} & \citep{zhu_neurips_21,churchland2010cortical} & \cmark & \cmark & & & \\
	21 & NeurIPS & Jensen \etal \citep{jensen_neurips_21} & \citep{o2017nonhuman} & & \cmark & & & \cmark \\
	21 & NeurIPS & Hurwitz \etal \citep{hurwitz_neurips_2021} & \citep{gallego_long-term_2020} & & \cmark & & & \cmark \\
	21 & NeurIPS & Bashiri \etal \citep{bashiri_neurips_2021} & \citep{bashiri_neurips_2021} & \cmark & \cmark & & & \cmark \\

	\bottomrule
    \end{tabular}
    \caption{LVMs for neural data that have been published in ML venues in the last 15 years. Neural LVMs have become increasingly frequent but without any increased standardization. A check under a metric column indicates what data the model inference is evaluated against.}
    \label{tbl:lvm_index}
    
\end{threeparttable}
}
\end{table}

As stated in our introduction, neural LVMs play an important role in analyzing the increasingly large neural datasets we can now record. We highlight in~\tableref{tbl:lvm_index} related works that have been published in ML venues, to demonstrate the role the neuro-LVM sub-community plays within broader ML dialogue. Additionally, the table shows the wide variety of datasets and evaluation strategies used in these works. Note that even when the same dataset or metric is listed, very few implementations are consistently applied. For example, subsets of datasets are taken, metrics are normalized to different baselines, and model inputs and outputs vary. This provides a semi-quantitative account of the need for the NLB.

\subsection{Datasets}

\subsubsection{Data format}

The datasets have been converted to the Neurodata Without Borders (NWB) format \cite{rubel2021neurodata}, a standardized format for neurophysiological data. Neurodata Without Borders provides open-source Python and Matlab APIs for reading NWB datasets, which can be found from its GitHub organization at \href{https://github.com/NeurodataWithoutBorders}{https://github.com/NeurodataWithoutBorders}. The data format is based on HDF5, allowing any programming language with an HDF5 package to access the data as a typical HDF5 file. In addition, for our benchmark, we provide a code package that facilitates reading from our converted NWB files and extracting relevant data, available at \href{https://github.com/neurallatents/nlb\_tools}{https://github.com/neurallatents/nlb\_tools}.

\subsubsection{Data hosting and licensing}

The datasets are hosted on the platform  DANDI (Distributed Archives for Neurophysiology Data Integration). DANDI is a platform specifically for publishing and sharing neurophysiology data. DANDI automatically generates structured metadata and persistent identifiers for the datasets uploaded to the site.
The datasets are distributed under a Creative Commons Attribution 4.0 International license. The authors bear all responsibility in case of violation of rights.
The datasets are available at the following links:
\begin{itemize}
    \item \maze - \href{https://dandiarchive.org/dandiset/000128}{https://dandiarchive.org/dandiset/000128}
    \item \rtt - \href{https://dandiarchive.org/dandiset/000129}{https://dandiarchive.org/dandiset/000129}
    \item \areatwo - \href{https://dandiarchive.org/dandiset/000127/}{https://dandiarchive.org/dandiset/000127/}
    \item \dmfc - \href{https://dandiarchive.org/dandiset/000130}{https://dandiarchive.org/dandiset/000130}
    \item \mazeL - \href{https://dandiarchive.org/dandiset/000138}{https://dandiarchive.org/dandiset/000138}
    \item \mazeM - \href{https://dandiarchive.org/dandiset/000139}{https://dandiarchive.org/dandiset/000139}
    \item \mazeS - \href{https://dandiarchive.org/dandiset/000140}{https://dandiarchive.org/dandiset/000140}
\end{itemize}

\subsubsection{Dataset documentation}

All datasets curated for this benchmark have been featured in previous works. In the Source section for each dataset, we point to the original papers where one can find more detailed technical information on the datasets. Because each dataset features recordings from rhesus macaques, there is no personally identifying information or offensive content in the datasets.

Since all our datasets feature neural spiking activity from macaques, they share certain processing steps. In particular, all were spike sorted. Spike sorting is the process of identifying individual neurons, or units, from the voltage readings recorded by extracellular electrodes, which can pick up activity from multiple neurons. Spike sorting is an imperfect process, and multiple electrodes may pick up the same neural activity and sort it into separate units. This serves as the motivation for our removal of some sorted units based on cross-correlation.

Below, we document the source, our intended use, experimental design, data collection, and the additional processing we applied  for each dataset.

\newpage
\subsubsubsection{\maze}

\begin{table}[htbp]
    \centering
    \resizebox{1.0\linewidth}{!}{
        \begin{tabular}{@{}cccccccc@{}}
            Name &
            Subject & 
            Session date & 
            Conditions &
            Training trials & 
            Test trials & 
            Held-in units & 
            Held-out units
            \\\toprule
\maze &
\multirow{4}{*}{\shortstack{Name: Jenkins\\Species: Macaca mulatta}} & 
2009-09-25 & 108 & 2295 & 574 & 137 & 45 \\
\mazeS & & 2009-09-28 & 27 & 100 & 100 & 107 & 35\\
\mazeM & & 2009-09-29 & 27 & 250 & 100 & 114 & 38\\
\mazeL & & 2009-10-06 & 27 & 500 & 100 & 122 & 40\\
            \bottomrule
        \end{tabular}
    }
    \vspace{5pt}
    \caption{\footnotesize\xhdr{\maze summary.} Overview of assets included in \maze datasets.}
\end{table}

\paragraph{General description.\hspace{-2mm}}
This dataset contains sorted unit spiking times and behavioral data from a macaque performing a delayed reaching task. The experimental task was a center-out reaching task with obstructing barriers forming a maze, resulting in a variety of straight and curved reaches. Neural activity was recorded from electrode arrays implanted in the motor cortex (M1) and dorsal premotor cortex (PMd). Cursor position, hand position, and eye position were also recorded during the experiment, and hand velocity was calculated offline from hand position.

\paragraph{Source.\hspace{-2mm}}
This dataset was collected by Matthew T. Kaufman and Mark M. Churchland from the Shenoy Lab at Stanford University. The dataset was created for the purpose of examining macaque neural activity during movement preparation and execution. The experiment and data collection is described in \cite{churchland2010cortical} and the dataset has been featured in a number of papers, including ~\cite{keshtkaran2019enabling,churchland2010cortical,Churchland2012-JPCA,Pandarinath2018-LFADS,hennequin2014optimal,elsayed2017structure,kaufman2010roles,sussillo2015neural,kaufman2013roles,kaufman2014cortical,kaufman2016largest,kaufman2015vacillation,ye2021representation,keshtkaran2021large}. The dataset creators have granted permission to use and distribute the dataset sessions as part of the benchmark.

\paragraph{Intended use.\hspace{-2mm}}
This dataset has been curated for use in evaluating latent variable models of neural spiking activity as part of the Neural Latents Benchmark. The dynamics of the motor and premotor cortices during movement preparation and execution have been found to be well-modeled by autonomous dynamical systems. As such, the dataset tests how well models can infer such dynamics from noisy spiking activity. The inclusion of multiple sessions with varying trial counts also serves to evaluate how model performance scales with the amount of training data. The dataset is available on the platform DANDI to allow others to evaluate their methods on the data.

\paragraph{Experiment design.\hspace{-2mm}}
The maze task is a delayed-reach task. The task was performed with a cursor that was projected slightly above the monkey's fingertips, which were tracked using a reflective bead. The task was performed in the plane of a vertical screen. The monkeys had to keep their hand close to the screen, but did not need to (and did not) slide along it.

In any particular trial, the monkey first fixated the eye and cursor inside a central fixation point. The monkey was then presented with a target and potentially barriers and distracters, which are unreachable targets that the monkey should ignore. The monkey maintained fixation until the go cue was signalled, indicated by the disappearance of the central fixation point and visual changes to the target. The monkey then made a rapid reach to the target and held at the target location for a short duration. If the cursor collided with a barrier before reaching the target, the display cleared and the trial ended.

The experimenters hand-designed 4 sets of 27 maze configurations. Each set of 27 maze configurations was grouped into 3 subsets of 3 mazes, each with 3 versions. The three versions of each maze have 1 target and no barriers, 1 target and barriers, or 1 target, 2 distracter targets, and barriers. Mazes within each subset were identical except for the placement of a few barriers directly obstructing the targets. In addition to hand-designed mazes, impossible mazes where no targets were initially reachable and randomly generated mazes were presented.

\paragraph{Data collection methods.\hspace{-2mm}}
Neural activity was recorded using two 96-electrode Utah arrays implanted in the primary motor (M1) and dorsal premotor (PMd) cortices. Recordings were sampled at 30 kHz and spike sorted offline. Sorted units were manually rated per trial by stability and cleanness.

Target and go cue presentation time were recorded using a simultaneous flash of light out of the monkeys’ field of view that was detected using a “photo box”. Monkey movement onset time and reaction time were calculated offline using a robust algorithm.

Hand and eye tracking were used to record the monkey’s movements and gaze. Cursor position was calculated real-time from hand position with various filtering and inertial prediction to make movement smoother. Values were recorded at 1 millisecond resolution.

\paragraph{Processing.\hspace{-2mm}}
For all sessions, trials with random or impossible mazes, unsuccessful reaches, or potential recording issues were discarded. Sorted units with low unit quality ratings or firing rates less than 0.1 Hz were removed. Cross-correlations for all pairs of sorted units were computed and neurons were removed until all cross-correlations were below a threshold. Thresholds were determined by plotting histograms of all cross-correlation values and identifying outliers. Neurons within each pair were removed based on which neuron was in more threshold-surpassing pairs, or, if that number was the same, by which had higher average cross-correlation with all other units. 25\% of the remaining sorted units were randomly selected to be held-out in test data. A recording error in hand position was corrected by subtracting 8 mm from all {\it y} measurements. Hand velocity was estimated from hand position using second order accurate central differences. For \maze, 20\% of the trials in each condition were randomly selected for the test set. For \mazeS, \mazeM, and \mazeL, trials were randomly selected from each condition for the train, val, and test sets. Ordering of trials in the train, val, and test sets was randomized for all sessions.

\newpage
\subsubsubsection{\rtt}

\begin{table}[htb]
    \centering
    \resizebox{1.0\linewidth}{!}{
        \begin{tabular}{@{}ccccccc@{}}
            Name &
            Subject & 
            Session date & 
            Training trials & 
            Test trials & 
            Held-in units & 
            Held-out units
            \\\toprule
\rtt & \shortstack{Name: Indy\\Species: Macaca mulatta} & 2017-07-02 & \shortstack{10.8 min\\1080 trials} & \shortstack{3.62 min\\271 trials} & 98 & 32\\
            \bottomrule
        \end{tabular}
    }
    \vspace{5pt}
    \caption{\footnotesize\xhdr{\rtt summary.} Overview of assets included in \rtt dataset.}
\end{table}

\paragraph{General description.\hspace{-2mm}}
This dataset contains sorted unit spiking times and behavioral data from a macaque performing a self-paced reaching task. In the experimental task, the subject reached between targets randomly selected from an 8x8 grid without gaps or pre-movement delay intervals. Neural activity was recorded from an electrode array implanted in the primary motor cortex. Finger position, cursor position, and target position were also recorded during the experiment.

\paragraph{Source.\hspace{-2mm}}
This dataset was collected by Joseph E. O’Doherty from the Sabes Lab at UCSF. The dataset was created for the purpose of examining neural activity during naturalistic self-paced motor behavior and evaluating the performance of behavioral decoding algorithms. Other sessions from this dataset and descriptions of the included data are available publicly on Zenodo \cite{o2017nonhuman}. The experimental procedure is described in detail in \cite{makin2018decoder}, from which the descriptions below are taken. The dataset creator has granted permission to use and distribute the dataset as part of the benchmark.

\paragraph{Intended use.\hspace{-2mm}}
This dataset has been prepared for use in evaluating latent variable models of neural spiking activity as part of the Neural Latents Benchmark. The unique task design results in the absence of pre-movement delay periods and repeated, highly constrained reach conditions typical of other reaching tasks. The ability to accurately model motor cortical activity during this more naturalistic behavior, and not only highly-stereotyped reaches with preparatory periods, is vital to developing effective BCI decoders and understanding the functioning of the motor cortex. The dataset is available on the platform DANDI to allow others to evaluate their methods on the data.

\paragraph{Experiment design.\hspace{-2mm}} 
In the experiment, circular targets were presented in an 8-by-8 square grid. The monkey made movements in the horizontal plane and was shown the targets in a virtual reality environment using a mirror and projector. The monkey acquired targets by placing the fingertip of the working arm within a square acceptance zone centered on each target for 450ms. Targets were spaced such that acceptance zones were non-overlapping. After target acquisition, a new target was randomly drawn with replacement from the set of possible target locations and presented immediately. For a period of 200 ms after target acquisition, the next target was ``locked-out'' and could not be acquired.

\paragraph{Data collection methods.\hspace{-2mm}}
Neural activity was recorded using a single 96-electrode Utah array implanted in the primary motor cortex. Recordings were made using an RZ2 BioAmp Processor with PZ2 Preamplifier (TuckerDavis Technologies, Alachua, FL). The recordings were sampled at 24.4 kHz and filtered with a causal IIR bandpass filter (fourth-order Butterworth; Fpass = 500 Hz to 5000 Hz). Spikes were detected and sorted online using custom software.

Fingertip position was monitored with a six-axis electromagnetic position sensor (Polhemus Liberty, Colchester, VT) at 250 Hz. After acquisition, the position data were non-causally low-pass filtered to reject sensor noise (4th order Butterworth; Fcutoff = 10 Hz).

\paragraph{Processing.\hspace{-2mm}}
Unsorted “hash” units and sorted units with firing rates less than 0.1 Hz were removed. Cross-correlations for all pairs of sorted units were computed and neurons were removed until all cross-correlations were below a threshold. Thresholds were determined by plotting histograms of all cross-correlation values and identifying outliers. Neurons within each pair were removed based on which neuron was in more threshold-surpassing pairs, or, if that number was the same, by which had higher average cross-correlation with all other units. 25\% of the remaining units were randomly selected to be held-out in test data. Finger velocity was estimated from finger position using second-order accurate central differences. The first 4.5 seconds and last 0.9 seconds of the continuous session were discarded as the subject was not actively performing reaches in those periods. The remaining continuous recording was divided into 9 segments alternating between test and train, resulting in 5 test segments and 4 train segments. The lengths of the test segments were randomly altered while maintaining roughly equal train segments. The sizes of the test segments were limited to ensure they would collectively contain roughly 20\% of the total “trials”, which were simply continuous 600 ms snippets of the recording. In the test set, an additional 200 ms of data was held-out per “trial” for forward prediction evaluation. Ordering of trials in the test set was randomized. The train segments were randomly divided into train and val splits.

\newpage
\subsubsubsection{\areatwo}

\begin{table}[h]
    \centering
    \resizebox{1.0\linewidth}{!}{
        \begin{tabular}{@{}cccccccc@{}}
            Name &
            Subject & 
            Session date & 
            Conditions & 
            Training trials & 
            Test trials & 
            Held-in units & 
            Held-out units
            \\\toprule
\areatwo & \shortstack{Name: Han\\Species: Macaca mulatta} & 2017-12-07 & 16 & 364 & 98 & 49 & 16\\
            \bottomrule
        \end{tabular}
    }
    \vspace{5pt}
    \caption{\footnotesize\xhdr{\areatwo summary.} Overview of assets included in \areatwo dataset.}
\end{table}

\paragraph{General description.\hspace{-2mm}}
This dataset contains sorted unit spiking times and behavioral data from a macaque performing a reaching task with perturbations. In the experimental task, the subject performed delayed center-out reaches using a manipulandum to control a cursor. On a portion of the trials, the manipulandum applied a bump during the center hold prior to the reach. Neural activity was recorded from an electrode array implanted in the somatosensory area 2. Hand position, cursor position, force applied to the manipulandum, length and velocity of various arm muscles, and angle and velocity of various arm joints were all recorded during the experiment.

\paragraph{Source.\hspace{-2mm}}
This dataset was collected by Raeed Chowdhury from the Miller Lab at Northwestern University. The dataset was collected for the purpose of examining neural encoding of proprioceptive information. The experiment and data collection are described in \cite{chowdhury2020area}, from which much of the information below was taken. The dataset creator has granted permission to use and distribute the dataset as part of the benchmark.

\paragraph{Intended use.\hspace{-2mm}}
This dataset has been prepared for use in evaluating latent variable models of neural spiking activity as part of the Neural Latents Benchmark. Area 2 is robustly driven by mechanical perturbations to the arm and has been shown to contain information about whole-arm kinematics. Thus, the experimental task results in both typical proprioceptive input and unexpected perturbations in the recorded brain area. As such, this dataset provides the challenge of modeling input-driven activity in response to both predictable and unpredictable sensory feedback. The dataset is available on the platform DANDI to allow others to evaluate their methods on the data.

\paragraph{Experiment design.\hspace{-2mm}}
In the experiment, the monkey performed a classic center-out reaching task. The monkey used a manipulandum to control a cursor to reach for targets presented on a screen in a 20 cm x 20 cm workspace. The experiment consisted of two types of trials: active and passive. In active trials, the monkey held the cursor in a target at the center of the workspace for a random amount of time, after which one of eight outer targets was presented. The monkey then reached toward the target, and the trial ended in success once the monkey reached the outer target. On passive trials, motors on the manipulandum delivered a 2 N perturbation to the monkey’s hand in one of the target directions during the center hold period. After the bump, the monkey returned to the center target, after which the trial proceeded like an active trial. Active and passive trials each made up 50\% of the total trials.

\paragraph{Data collection methods.\hspace{-2mm}}
Neural activity was recorded from a 96-electrode Utah array implanted in the arm representation of area 2 of somatosensory cortex.  The Cerebus recording system (Blackrock) was used to record neural data for the experiments. Signals were sampled at 30 kHz and threshold crossings were detected online. After data collection, the Plexon Offline Sorter was used to manually sort threshold crossings into putative single units.

The position of the handle  was recorded using encoders on the manipulandum joints. The interaction forces between the monkey’s hand and the handle were recorded  using a six-axis load cell mounted underneath the handle. Before each reaching experiment, markers of different colors were painted on the outside of the monkey’s arm. A custom motion tracking system built from a Microsoft Kinect was used to record the 3D locations of these markers with respect to the camera, synced in time to the other behavioral and neural data. Muscle length and velocity and joint angle and velocity were computed from the motion tracking data.

\paragraph{Processing.\hspace{-2mm}}
Sorted units with firing rates less than 0.1 Hz were removed. Cross-correlations for all pairs of sorted units were computed and neurons were removed until all cross-correlations were below a threshold. Thresholds were determined by plotting histograms of all cross-correlation values and identifying outliers. Neurons within each pair were removed based on which neuron was in more threshold-surpassing pairs, or, if that number was the same, by which had higher average cross-correlation with all other units. The continuous recording was divided into 9 segments alternating between test and train, resulting in 5 test segments and 4 train segments. While the number of trials in train segments were kept roughly equal, the number of trials in test segments were randomly selected from the set of values that ensured there would be at least 4 trials of each condition in the test data. The total number of test trials was held to around 20\% of the number of successful trials. Ordering of trials in the test set was randomized. The train segments were randomly divided into train and val splits.

\newpage
\subsubsubsection{\dmfc}

\begin{table}[h]
    \centering
    \resizebox{1.0\linewidth}{!}{
        \begin{tabular}{@{}cccccccc@{}}
            Name &
            Subject & 
            Session date & 
            Conditions & 
            Training trials & 
            Test trials & 
            Held-in units & 
            Held-out units
            \\\toprule
\dmfc & \shortstack{Name: Haydn\\Species: Macaca mulatta} & 2016-12-11 & 40 & 1006 & 283 & 40 & 14\\
            \bottomrule
        \end{tabular}
    }
    \vspace{5pt}
    \caption{\footnotesize\xhdr{\dmfc summary.} Overview of assets included in \dmfc dataset.}
\end{table}

\paragraph{General description.\hspace{-2mm}}
This dataset contains sorted unit spiking times from a macaque performing a time-interval reproduction task. In the experimental task, the monkey was presented with two stimuli separated by a specific interval of time. The monkey then attempted to time their response such that the interval between the second stimulus and their response matched the interval separating the two stimuli. Neural activity was recorded from neural probes implanted in the dorsomedial frontal cortex.

\paragraph{Source.\hspace{-2mm}}
This dataset was collected by Hansem Sohn from the Jazayeri Lab at MIT. The dataset was collected for the purpose of examining the neural computations underlying Bayesian inference. The experiment and data collection are described in \cite{sohn2019bayesian}, from which much of the information below was taken. The dataset creator has granted permission to use and distribute the dataset session as part of the benchmark.

\paragraph{Intended use.\hspace{-2mm}}
This dataset has been prepared for use in evaluating latent variable models of neural spiking activity as part of the Neural Latents Benchmark. Recorded from a cognitive brain area, the dataset poses the challenge of modeling complex neural activity without a clear moment-by-moment behavioral correlate. The dataset is available on the platform DANDI to allow others to evaluate their methods on the data.

\paragraph{Experiment design.\hspace{-2mm}}
The Ready-Set-Go (RSG) task is a time-interval reproduction task. At the beginning of each trial, the monkey was presented with two fixation cues: a circle, indicating visual fixation on the display center, and a square, instructing the monkey to hold the joystick in its central position. While fixating, two visual flashes – Ready followed by Set – provided the first two beats of an isochronous rhythm. The monkey estimated the sample interval, $t_s$, between Ready and Set (i.e., estimation epoch), and used this information in the subsequent production epoch to generate the omitted third beat (Go) by initiating a response action. Monkeys received reward if the produced interval, $t_p$, between Set and Go was sufficiently close to $t_s$. The Go response action was an eye saccade or joystick movement to the right or left. The response modality was indicated by the color of the fixation cues. The response direction was indicated by the placement of the displayed target.

The sample interval $t_s$ was sampled from one of two prior distributions, a ‘Short’ prior ranging between 480 and 800 ms, and a ‘Long’ prior ranging between 800 and 1200 ms. The full experiment consisted of 40 conditions: 5 $t_s$ values each for the two priors (‘Short’ and ‘Long’), two response modalities (eye saccade and joystick movement), and two target directions. Like the response modality, the prior condition was cued explicitly by the color of the fixation cues. Response modality and prior condition were varied in blocks of trials. The target direction was chosen randomly across trials.

\paragraph{Data collection methods.\hspace{-2mm}}
Neural activity was recorded with 3 Plexon probes implanted in the dorsomedial frontal cortex. Signals were amplified, bandpass filtered, sampled at 30 kHz, and saved using the CerePlex data acquisition system (Blackrock Microsystems, UT). Spikes from single-units and multi-units were sorted offline using Kilosort software suites.

\paragraph{Processing.\hspace{-2mm}}
Due to recording instabilities, many sorted units had substantial changes in firing rate through the session. Firing rates for each unit were plotted and visually inspected, and units that dropped out for portions of the session were removed. The beginning and end of the session were also removed, as an excessive number of units showed instabilities in those periods. Remaining sorted units with firing rates less than 0.1 Hz were removed. Cross-correlations for all pairs of remaining sorted units were computed and neurons were removed until all cross-correlations were below a threshold. Thresholds were determined by plotting histograms of all cross-correlation values and identifying outliers. Neurons within each pair were removed based on which neuron was in more threshold-surpassing pairs, or, if that number was the same, by which had higher average cross-correlation with all other units. The continuous recording was divided into 13 segments alternating between test and train, resulting in 7 test segments and 6 train segments. While the number of trials in train segments were kept roughly equal, the number of trials in test segments were randomly selected from the set of values that ensured there would be at least 4 trials of each condition in the test data. The total number of test trials was held to around 20\% of the number of successful trials. Ordering of trials in the test set was randomized. The train segments were randomly divided into train and val splits.

\newpage
\subsection{Baselines}

To seed our benchmark, we applied 5 baseline methods of varying complexity to all 7 datasets. Baselines were run 3 times with different random seeds, and mean scores $\pm$ standard error of the mean were reported.. Spike smoothing, which simply fits a GLM from smoothed spikes to held-out spiking activity, consistently converges on the same solution regardless of random initialization and thus was not run multiple times. Code for our implementations of each method are available in our GitHub repo: \href{https://github.com/neurallatents/nlb\_tools}{https://github.com/neurallatents/nlb\_tools}. The scripts used for each baseline, their dependencies, and parameter search ranges can be found in the \texttt{examples/baselines/} directory. We summarize our approaches below. Note that reported values for computation resources used are only for one baseline run, not all three.

\subsubsection*{Spike smoothing}

\paragraph{Implementation.\hspace{-2mm}}
Held-in spiking activity was convolved with a Gaussian kernel. A Poisson GLM was fit from the logarithm of the smoothed spikes (with a small offset to prevent taking the log of 0) to held-out spiking activity for co-smoothing rate predictions.

\paragraph{Parameter optimization.\hspace{-2mm}}
Gaussian kernel standard deviation and Poisson GLM regularization penalty were optimized with a grid search. Models were validated with 5-fold cross-validation within the training data. The best set of parameters was used to train a new model on the entire training set and make predictions on the test set.

\begin{table}[htb]
    \centering
        \begin{tabular}{@{}|c|c|c|c|@{}}
            Dataset &
            GPU count & 
            GPU type & 
            Runtime
            \\\toprule
\maze & 0 & NA & 3.2 hrs\\
\mazeS & 0 & NA & 0.19 hrs\\
\mazeM & 0 & NA & 0.42 hrs\\
\mazeL & 0 & NA & 0.88 hrs\\
\rtt & 0 & NA & 1.67 hrs\\
\areatwo & 0 & NA & 0.30 hrs\\
\dmfc & 0 & NA & 0.62 hrs\\
            \bottomrule
        \end{tabular}
    \vspace{5pt}
    \caption{\footnotesize\xhdr{Spike smoothing computational resources.} Resources used for spike smoothing parameter optimization.}
\end{table}

\paragraph{Code availability.\hspace{-2mm}}
Spike smoothing was implemented using the standard Python libraries numpy, scipy, and scikit-learn.

\subsubsection*{GPFA}

\paragraph{Implementation.\hspace{-2mm}}
GPFA was run using the elephant Python package. The fit GPFA model was then used to transform test held-in spiking data to latent factors. Linear regression was fit from latent factors to held-in rates. Non-positive held-in rates were rectified to a small positive value. A Poisson GLM was fit from held-in rate predictions to held-out spikes to generate held-out rate predictions.

\paragraph{Parameter optimization.\hspace{-2mm}}
Latent dimensionality and regularization penalties for both the linear regression and Poisson GLM were optimized through grid search. Models were validated with 3-fold cross-validation within the training data. The best set of parameters was used to train a new model on the entire training set and make predictions on the test set.

\begin{table}[htb]
    \centering
        \begin{tabular}{@{}|c|c|c|c|@{}}
            Dataset &
            GPU count & 
            GPU type & 
            Runtime
            \\\toprule
\maze & 0 & NA & 35.7 hrs\\
\mazeS & 0 & NA & 1.25 hrs\\
\mazeM & 0 & NA & 3.0 hrs\\
\mazeL & 0 & NA & 8.8 hrs\\
\rtt & 0 & NA & 9.44 hrs\\
\areatwo & 0 & NA & 3.3 hrs\\
\dmfc & 0 & NA & 19.2 hrs\\
            \bottomrule
        \end{tabular}
    \vspace{5pt}
    \caption{\footnotesize\xhdr{GPFA computational resources.} Resources used for GPFA parameter optimization.}
\end{table}

\paragraph{Code availability.\hspace{-2mm}}
In addition to standard Python libraries, our GPFA implementation uses the public package elephant, available at \href{https://github.com/NeuralEnsemble/elephant}{https://github.com/NeuralEnsemble/elephant}.

\subsubsection*{SLDS}

\paragraph{Implementation.\hspace{-2mm}}
We used a modified version of the Linderman Lab's ssm package for our implementation of SLDS. SLDS was trained on spiking activity from all neurons for both required and forward prediction time steps. To generate test predictions, we approximated the posterior on held-in test spiking activity, using masks to indicate the missing held-out test data. Using the approximated posteriors, rate predictions were generated by smoothing observations for both the train and test data.

\paragraph{Parameter optimization.\hspace{-2mm}}
Latent dimensionality, number of discrete states, and the dynamics regularization penalty were optimized with a random search. Due to the long training times of SLDS, especially on datasets with large numbers of trials, parameter combinations were trained and evaluated on two random samples of 100 training and 100 validation trials (or 50 training and 50 validation for the \mazeS). Random searches were performed with 20 parameter combinations. Models were trained for 50 iterations during the random search. After the random search, the three parameter combinations with the best co-smoothing scores were trained on the full training data for 50 and 100 iterations. The best results out of these 6 models were selected. If performance was unsatisfactory after the random search, parameters were further hand-tuned based on trends in random search results. Due to our time constraints, it is likely that SLDS performance can still be improved over our results.

\begin{table}[htb]
    \centering
        \begin{tabular}{@{}|c|c|c|c|@{}}
            Dataset &
            GPU count & 
            GPU type & 
            Runtime
            \\\toprule
\maze & 0 & NA & 73.2 hrs\\
\mazeS & 0 & NA & 8.2 hrs\\
\mazeM & 0 & NA & 11.8 hrs\\
\mazeL & 0 & NA & 17.2 hrs\\
\rtt & 0 & NA & 19.3 hrs\\
\areatwo & 0 & NA & 12.4 hrs\\
\dmfc & 0 & NA & 40.1 hrs\\
            \bottomrule
        \end{tabular}
    \vspace{5pt}
    \caption{\footnotesize\xhdr{SLDS computational resources.} Resources used for SLDS parameter optimization.}
\end{table}

\paragraph{Code availability.\hspace{-2mm}}
Our modified version of ssm is available on GitHub: \href{https://github.com/felixp8/ssm}{https://github.com/felixp8/ssm}. In the modified package, parts of ssm were re-implemented using PyTorch to reduce runtime. The original ssm package is available on GitHub as well: \href{https://github.com/lindermanlab/ssm}{https://github.com/lindermanlab/ssm}.

\subsubsection*{AutoLFADS}

\paragraph{Implementation.\hspace{-2mm}}
AutoLFADS was run using SNEL's internal Python implementation. The model architecture was modified to only take held-in spiking activity as input to the encoder while outputting held-in and held-out rate predictions (for both held-out units and held-out timepoints).

\paragraph{Parameter optimization.\hspace{-2mm}}
Parameter optimization is done in AutoLFADS through Population Based Training.

\begin{table}[htb]
    \centering
        \begin{tabular}{@{}|c|c|c|c|@{}}
            Dataset &
            GPU count & 
            GPU type & 
            Runtime
            \\\toprule
\maze & 10 & Nvidia GeForce RTX 2080 & 4.66 hrs\\
\mazeS & 10 & Nvidia GeForce RTX 2080 & 0.58 hrs\\
\mazeM & 10 & Nvidia GeForce RTX 2080 & 0.97 hrs\\
\mazeL & 10 & Nvidia GeForce RTX 2080 & 1.94 hrs\\
\rtt & 10 & Nvidia GeForce RTX 2080 & 1.5 hrs\\
\areatwo & 10 & Nvidia GeForce RTX 2080 & 0.49 hrs\\
\dmfc & 10 & Nvidia GeForce RTX 2080 & 1.89 hrs\\
            \bottomrule
        \end{tabular}
    \vspace{5pt}
    \caption{\footnotesize\xhdr{AutoLFADS computational resources.} Resources used for AutoLFADS parameter optimization.}
\end{table}

\paragraph{Code availability.\hspace{-2mm}}
Because \cite{keshtkaran2021large} is still under review, SNEL will not publicly release the code right now. The code will be made available when the paper is published. A separate implementation of AutoLFADS is available at \href{https://github.com/snel-repo/autolfads}{https://github.com/snel-repo/autolfads}. However, this implementation does not contain the modifications applied for co-smoothing.

\subsubsection*{NDT}

\paragraph{Implementation.\hspace{-2mm}}
NDT was run with the public implementation linked below. The model architecture was modified to treat held-out neurons and timesteps as additional masked elements to predict in every sample.

\paragraph{Parameter optimization.\hspace{-2mm}}
Parameter optimization is through random search in a predefined grid (see public config files).

\begin{table}[htb]
    \centering
        \begin{tabular}{@{}|c|c|c|c|@{}}
            Dataset &
            GPU count & 
            GPU type & 
            Runtime
            \\\toprule
\maze & 10 & Nvidia GeForce RTX 2080 & 4.5 hrs\\
\mazeS & 10 & Nvidia GeForce RTX 2080 & 0.5 hrs\\
\mazeM & 10 & Nvidia GeForce RTX 2080 & 1.0 hrs\\
\mazeL & 10 & Nvidia GeForce RTX 2080 & 2.0 hrs\\
\rtt & 10 & Nvidia GeForce RTX 2080 & 1.5 hrs\\
\areatwo & 10 & Nvidia GeForce RTX 2080 & 0.5 hrs\\
\dmfc & 10 & Nvidia GeForce RTX 2080 & 2.0 hrs\\
            \bottomrule
        \end{tabular}
    \vspace{5pt}
    \caption{\footnotesize\xhdr{NDT computational resources.} Resources used for NDT parameter optimization.}
\end{table}

\paragraph{Code availability.\hspace{-2mm}}
The NDT code is publicly available at \href{https://github.com/snel-repo/neural-data-transformers}{https://github.com/snel-repo/neural-data-transformers}.

\subsection{Evaluation Parameters}

A number of parameters are used in evaluation, namely trial alignment ranges, decoding lags, and PSTH kernel standard deviations. Below, we describe our processes for choosing these values for each dataset.

\subsubsection{Trial alignment}

For \maze, \mazeL, \mazeM, and \mazeS, trials are aligned from 250 ms before to 450 ms after movement onset. This alignment was chosen such that each trial contained a portion of both the pre-movement preparatory period and the actual reach.

For \rtt, trials are created by splitting the continuous data into 600 ms segments. This value was simply chosen as a fairly typical trial length. The ability to infer latents from segments of comparable or shorter length is essential for real-time decoding of neural signals.

For \areatwo, trials are aligned from 100 ms before to 500 ms after movement onset, either after target onset for active trials or in response to the bump for passive trials. This alignment was chosen so that each trial contained the majority of the corrective movement or reach. For passive trials, the alignment interval also guarantees the inclusion of the bump within the interval.

For \dmfc, the trials are aligned from 1500 ms before up to the go response, with up to 300 ms of jitter added to the alignment point. This alignment was chosen to include the entire production epoch (between Set and Go) for all successful trials in all conditions. While the estimation epoch (between Ready and Set) is also of interest, the combined length of the estimation and production epochs on the longest trials exceeds the length of the shortest trials. Taking only the production epoch guarantees that no segments will contain data from more than one trial.

\subsubsection{Decoding lag}
Because signals take time to travel between the brain and peripheral nerves, external behavior is expected to lag behind the brain activity which drives it. Thus, to find the optimal time difference between neural activity and behavior for decoding, we searched across lag amounts by evaluating decoding with 5-fold cross validation. We ran this evaluation for all 5 baseline methods, choosing the lag value giving the highest mean result across all methods. The resulting values are:
\begin{itemize}
    \item \maze: 100 ms
    \item \rtt: 140 ms
    \item \areatwo: -20 ms
    \item \mazeL: 120 ms
    \item \mazeM: 120 ms
    \item \mazeS: 120 ms
\end{itemize}
Note that \areatwo is recorded from a sensory area, so behavior is expected to precede the corresponding neural activity, which is confirmed by the negative optimal lag value.

\subsubsection{PSTH kernel width}
PSTH calculation typically involves two steps: convolution with a Gaussian kernel and averaging across trials. In order to choose a kernel standard deviation, we evaluated various kernel standard deviations using leave-one-out cross validation: For each trial, we computed PSTHs from all other trials in the same condition as the given trial. We then calculated the Poisson likelihood given that trial's spiking activity, using the PSTHs as the firing rates. The kernel standard deviation giving the highest mean likelihood across all trials was chosen for the PSTH metric. For \mazeL, though the optimal value was 40 ms, we instead use 50 ms in order to facilitate comparison with \mazeM and \mazeS, both of which have optimal values of 50 ms. The resulting PSTH kernel standard deviations are:
\begin{itemize}
    \item \maze: 70 ms
    \item \areatwo: 40 ms
    \item \dmfc: 70 ms
    \item \mazeL: 50 ms
    \item \mazeM: 50 ms
    \item \mazeS: 50 ms
\end{itemize}

\subsection{\dmfc Changes After Initial Release}

During the NeurIPS rebuttal period, LFADS random searches on \dmfc revealed models that achieved near-perfect correlation scores while performing extremely poorly on co-smoothing. These models also scored better on match to PSTH than models that performed very well on co-smoothing. After a series of new analyses, we determined the sources of the issues and subsequently made changes to our evaluation methods, data preparation, and dataset.

First, we concluded that the original PSTH metric implementation, which evaluated each individual trial's match to the empirical PSTHs, was poorly suited to datasets with high single-trial variance in neural responses. For such datasets, like \dmfc, excellent match to the neural data is detrimental to performance on the PSTH metric. Thus, we altered our PSTH metric to instead average model rate predictions within each condition and compare those to the empirical PSTHs. The original implementation penalizes accurate representation of single-trial variance that deviates from PSTHs, while the new implementation, by averaging across trials, mitigates this effect.

Second, we determined that our original trial windows aligned to the go response were problematic. Instead of using the entire trial window, average neural speed is calculated from only the set-go window, the length of which is $t_p$, one of the variables in our correlation calculation. Our original trial alignment guaranteed that the go response occurred precisely at end of the trial window, which we found allowed a trivial high correlation score. Specifically,
a model can output identical rate inferences in every trial, that increase in instantaneous neural speeds towards the end of the trial. The problematic alignment then includes more low-speed timesteps for longer trials, resulting in a lower average neural speed and a trivial correlation between neural speed and $t_p$. To remedy this, we introduced jitter to the alignment windows so that the go response does not always land exactly at the end of the trial window.

Finally, we found that a particular neuron dominated neural speed estimates. This neuron alone was sufficient to achieve excellent correlation scores, while its removal resulted in extremely poor scores for all models. This imbalance led us to change the dataset entirely for one recorded from a different session.

}{}

\end{document}